\documentclass{article}
\usepackage{microtype}
\usepackage{graphicx}
\usepackage{twemojis}
\usepackage{hyperref}
\usepackage[preprint]{icml2027}

\usepackage{booktabs}
\usepackage{multirow}
\usepackage{array}
\usepackage{amsmath}
\usepackage{amssymb}
\usepackage{amsthm}
\usepackage{enumitem}
\usepackage{graphicx}
\usepackage{xcolor}
\usepackage[most]{tcolorbox}
\usepackage{etoolbox}

\providecommand{\linenumbers}{}
\providecommand{\nolinenumbers}{}
\AtBeginEnvironment{figure*}{\nolinenumbers}
\AtBeginEnvironment{table*}{\nolinenumbers}
\AtBeginEnvironment{algorithm}{\nolinenumbers}

\newtheorem{theorem}{Theorem}
\newtheorem{proposition}{Proposition}
\newtheorem{corollary}{Corollary}[theorem]

\newtheorem{lemma}{Lemma}[section]
\theoremstyle{definition}
\newtheorem{assumption}{Assumption}

\newtcolorbox{promptbox}[1][]{enhanced, colback=gray!4, colframe=gray!55,
  boxrule=0.5pt, arc=2pt, left=7pt, right=7pt, top=3pt, bottom=5pt,
  title={#1}, fonttitle=\bfseries\small, coltitle=black,
  colbacktitle=gray!18, titlerule=0pt, toptitle=2pt, bottomtitle=2pt,
  before skip=6pt, after skip=6pt}
\newcommand{\MODEL}{\textsc{AGORA}}

\newcommand{\rev}[1]{#1}
\newcommand{\pol}[1]{#1}
\newcommand{\opt}[1]{#1}

\newcommand{\revred}[1]{#1}
\newcommand{\revp}[1]{#1}

\newcommand{\CV}{\mathrm{CV}}
\newcommand{\NSR}{\mathrm{NSR}}

\icmltitlerunning{\MODEL: Sample-Efficient Aggregation of LLM Judges with Heterogeneous Reliability}

\begin{document}

\twocolumn[
\icmltitle{\MODEL: Sample-Efficient Aggregation of LLM Judges\\with Heterogeneous Reliability}
\begin{icmlauthorlist}
\icmlauthor{Junhyung Ahn}{uiuc}
\end{icmlauthorlist}
\icmlaffiliation{uiuc}{School of Information Sciences, University of Illinois Urbana-Champaign, USA}
\icmlcorrespondingauthor{Junhyung Ahn}{ja86@illinois.edu}
\icmlkeywords{LLM-as-a-judge, item response theory, label aggregation}
\vskip 0.3in
]
\printAffiliationsAndNotice{}

\begin{abstract}
Large language models (LLMs) are widely used as judges to evaluate generative systems, yet their reliability, defined as the ability to distinguish differences in item quality, varies substantially across judges and evaluation dimensions. Aggregating their ratings is particularly challenging when only a limited number of items are available and human labels are absent. Existing methods either ignore differences in judge reliability, as in mean aggregation, or require estimating many parameters from limited data, as in Dawid--Skene. We propose AGORA (Aggregation of Reliability-weighted Assessments), an unsupervised method that jointly estimates latent item qualities and dimension-specific judge reliabilities from observed ratings alone. Grounded in item response theory, AGORA exploits the ordinal structure of rating scales to reduce the number of parameters per judge from quadratic to linear in the number of rating categories, while leveraging correlations across evaluation dimensions. Theoretically, we show that AGORA achieves greater accuracy gains over mean aggregation with fewer items as judges vary more widely in reliability, and is more sample-efficient than Dawid--Skene even when both models are correctly specified. Across three benchmarks, AGORA achieves the highest correlation with human ratings among the evaluated score aggregators at both small and full sample sizes, while its pairwise extension matches or outperforms existing baselines.\end{abstract}

\section{Introduction}
\label{sec:intro}

LLM judges have become a standard tool for evaluating generative systems at scale \citep{zheng2023judging,liu2023geval,verga2024juries}.
Individual judges show systematic biases \citep{zheng2023judging,wang2024fair}, so a common remedy is to score each item with several judges and to aggregate their scores.
This aggregation faces two problems: human labels are scarce, and the judges differ in reliability.

First, evaluation sets with human labels are small: Topical-Chat has $360$ dialogue responses \citep{mehri2020usr} and SummEval has $1{,}600$ summaries \citep{fabbri2021summeval}.
Because the performance of LLM judges can be measured only against human labels, aggregators are evaluated on these small labeled sets.
An aggregator must therefore estimate the reliability of each judge from the ratings of a few hundred items, without using the labels.

Second, the judges differ in reliability, which we define as how well a judge's ratings distinguish differences in item quality.
Reliability varies sharply across models and across the evaluation dimensions, such as coherence or relevance.
The most reliable judge on one dimension of a task can be among the least reliable on another \citep{bavaresco2025judgebench,nimah2023checklist,li2025jury}.
If all judges were equally reliable, the per-item mean aggregator would suffice, but when reliability varies widely across judges and dimensions, an accurate aggregate must weight the judges differently on each dimension.

Standard aggregators address only one of the two problems.
The mean does not address the second problem, since it gives every judge the same weight.
Annotation models such as Dawid--Skene \citep{dawid1979} learn unequal weights, but they do not address the first problem.
On a $K$-point scale, Dawid--Skene estimates a $K\times K$ table of rating probabilities for each judge, so a few hundred items leave each entry with only a few observations.

\begin{figure*}[t]
\centering
\includegraphics[width=\textwidth]{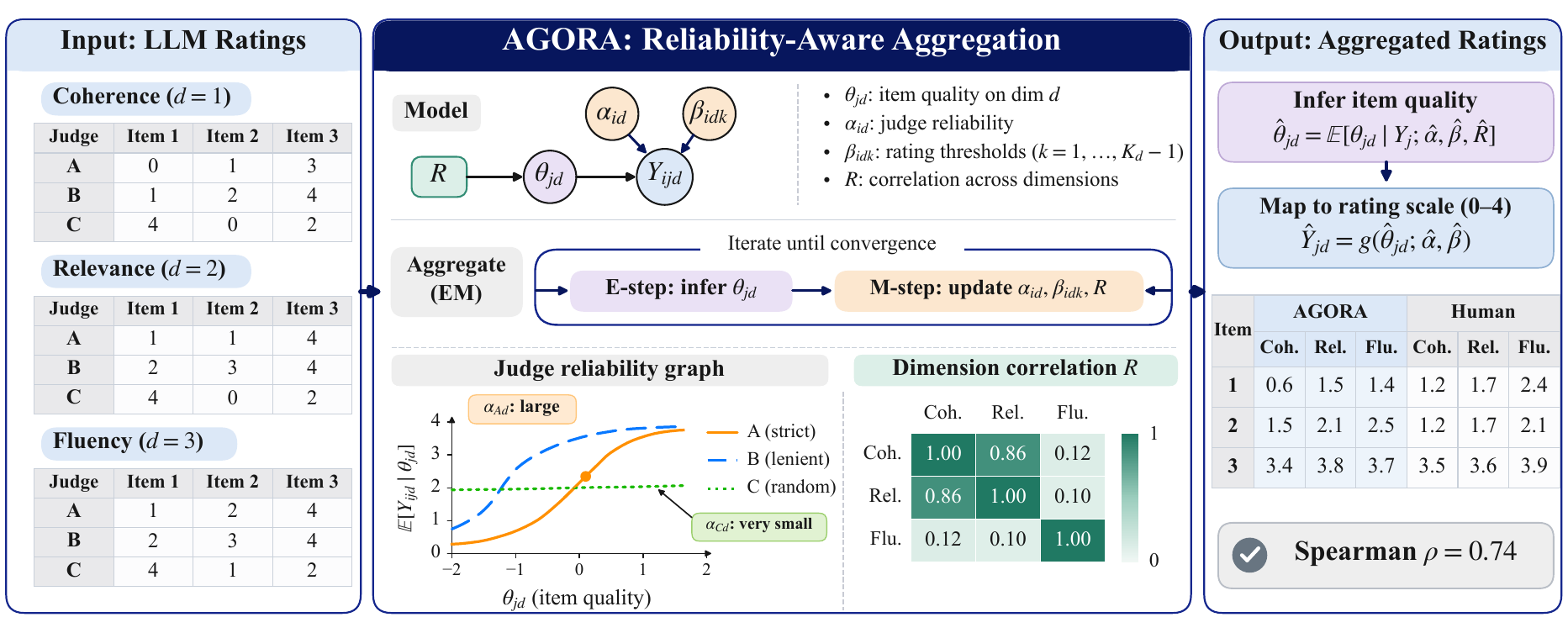}
\caption{Overview of \MODEL{}.
(Left) LLM judges rate each item on several dimensions.
(Middle) \MODEL{} models each rating $Y_{ijd}$ by the latent quality $\theta_{jd}$ of the item, the reliability $\alpha_{id}$ and the thresholds $\beta_{idk}$ of the judge, and the correlation $R$ across dimensions.
It alternates between inferring $\theta_{jd}$ and updating the other parameters, so an unreliable judge such as C gets a small $\alpha_{id}$.
(Right) \MODEL{} maps the estimated quality back to the rating scale, and we compare the result with human ratings.
The numbers are illustrative.}
\label{fig:example}
\end{figure*}

We propose \MODEL{}: \textbf{AG}gregation \textbf{O}f \textbf{R}eliability-weighted \textbf{A}ssessments, an aggregator that estimates the latent quality of each item on each evaluation dimension from the judges' ratings, weighting each judge by its estimated reliability.
Judge reliability measures how well a judge's ratings distinguish differences in item quality, while latent item quality is the underlying quality inferred from the ratings rather than directly observed.
\MODEL{} estimates both with the graded-response model of item response theory \citep{samejima1969}, which describes how each judge assigns ratings at different quality levels.
We fit the model with an expectation--maximization procedure that alternates between inferring the latent qualities and updating the judge parameters from the observed ratings.
The resulting estimates give greater influence to ratings that provide more information about item quality, so the random judge C in Figure~\ref{fig:example} has almost no effect.
Moreover, since evaluation dimensions can be related, \MODEL{} also learns a correlation matrix between their latent qualities, which lets ratings on related dimensions inform each quality estimate when the evaluation set is small.

We also prove that the accuracy gain of \MODEL{} over the mean increases with the reliability variation of the judges, and that the number of items required for this gain decreases accordingly.
Moreover, even when both models are correct, Dawid--Skene requires several times as many items as \MODEL{} to attain the same accuracy, because \MODEL{} treats the rating scale as ordered and therefore has far fewer parameters to estimate from the same items.

Our contributions are as follows:
\begin{enumerate}[leftmargin=*,itemsep=1pt,topsep=2pt]
  \item We propose \MODEL{}, an aggregator that learns the reliability of each judge from the ratings, weights the judges accordingly, and uses information across related dimensions (\S\ref{sec:method}).
  \item We prove that the accuracy gain of \MODEL{} over the mean increases with the reliability variation of the judges, and that Dawid--Skene needs several times as many items to match \MODEL{} even when both models are correct (\S\ref{sec:theory}).
  \item On three benchmarks, \MODEL{} has the highest correlation with human ratings among standard aggregators at both small and full size, and its pairwise extension matches or exceeds the pairwise baselines (\S\ref{sec:eval}).
\end{enumerate}

\section{Related Work}
\label{sec:relate}

\paragraph{Aggregating LLM judges.}
Most LLM-judge ensembles average or majority-vote their members \citep{verga2024juries,wang2024moa}, and some weight judges by a learned or heuristic reliability score \citep{guha2024smoothie}.
SkillAggregation \citep{sun2025skillaggregation} trains a neural encoder over binary judge probabilities, CARE decomposes the rating matrix into sparse and low-rank parts \citep{zhao2026care}, and \citet{ising2026} model judge dependence with Ising models.
\citet{btsigma2026} and \citet{xu2026judgeaware} add a per-judge discrimination to a Bradley--Terry model over pairwise comparisons.

\paragraph{Annotation models.}
Dawid--Skene, MACE, and related Bayesian annotation models \citep{dawid1979,hovy2013mace,paun2018comparing,demartini2012zencrowd} were developed for large crowdsourcing corpora, where each item receives many ratings \citep{zheng2017truth}.
GLAD \citep{whitehill2009glad} replaces the confusion matrix with an ability and difficulty model.

\paragraph{Item response theory.}
Item response theory (IRT) estimates the ability of each test taker from their answers to test questions \citep{lord1980,rasch1960}.
Its graded-response model handles answers on an ordinal scale, such as 1--5 ratings \citep{samejima1969}.
\citet{choi2026irtjudge} use this model to assess the reliability of a single LLM judge across prompt variants.
\MODEL{} uses it instead to combine the ratings of several judges into one score per item, weighting each judge by a reliability estimated without human labels.
Rater models in educational measurement also combine the ordinal scores of several raters \citep{linacre1989many,johnson1996multirater,patz2002hrm}.

\section{Problem Setting}
\label{sec:setting}

A benchmark rates items $j=1,\ldots,N$, such as summaries or translations, on dimensions $d=1,\ldots,D$, such as coherence and relevance.
A pool of judges $i=1,\ldots,J$ rates every item on every dimension.
Let $Y_d$ be the $J\times N$ rating matrix of dimension $d$, in which row $i$ holds the ratings of judge $i$ and column $j$ holds the ratings of item $j$.
Its entry $Y_{ijd}\in\{0,\ldots,K_d-1\}$ is an ordinal rating, where $K_d$ is the number of rating categories of dimension $d$.
The aggregator receives the rating matrices $Y=(Y_1,\ldots,Y_D)$ and returns one score per item and dimension.
We evaluate the scores by their Spearman correlation $\rho$ with the human ratings of the benchmark.

\section{The \MODEL{} Aggregator}
\label{sec:method}
This section describes how \MODEL{} estimates its parameters from the ratings and how it aggregates the ratings into a single score per item.
\S\ref{sec:base} and \S\ref{sec:borrow} define the parameters of the model, \S\ref{sec:estim} estimates these parameters from the ratings, and \S\ref{sec:readout} uses them to score each item.
Algorithm~\ref{alg:agora} summarizes the procedure (App.~\ref{app:algorithm}).

\subsection{Rating model of one dimension}
\label{sec:base}

{This subsection defines how a judge rates one dimension of an item, through a latent quality of the item and two parameters of the judge.
Let $\theta_{jd}\in\mathbb{R}$ be the latent quality of item $j$ on dimension $d$, where a higher value means a better item.
The ratings of different judges are independent given $\theta_{jd}$.
\MODEL{} models the rating of judge $i$ with the graded-response model \citep{samejima1969}:}
\begin{equation}
P(Y_{ijd}\ge k\mid\theta_{jd})
=
\sigma\big(\alpha_{id}\theta_{jd}-\beta_{idk}\big),
\label{eq:cum}
\end{equation}
for $k=1,\ldots,K_d-1$, where $\sigma$ is the logistic function.
The \emph{discrimination} $\alpha_{id}>0$ is the reliability of the judge: its ratings rise sharply with $\theta_{jd}$ when $\alpha_{id}$ is large and barely depend on it when $\alpha_{id}$ is near zero.
The ordered \emph{thresholds} $\beta_{id1}<\cdots<\beta_{id,K_d-1}$ describe how the judge uses the scale; a lenient judge, for example, has low thresholds.
Because each judge has its own thresholds, the same raw rating from two judges can indicate different latent qualities.
Each category has the probability
\begin{equation}
\begin{aligned}
P(Y_{ijd}=k\mid\theta_{jd})
={}&\sigma(\alpha_{id}\theta_{jd}-\beta_{idk})\\
&-\sigma(\alpha_{id}\theta_{jd}-\beta_{id,k+1}),
\end{aligned}
\label{eq:cat}
\end{equation}
with $\beta_{id0}=-\infty$ and $\beta_{idK_d}=+\infty$.

\subsection{Correlation across dimensions}
\label{sec:borrow}

The rating model treats each dimension separately, which ignores the other dimensions.
However, the dimensions of an item are related: an incoherent summary is often also less relevant~\citep{xiao2023metriceval}.
The ratings on one dimension therefore carry evidence about the latent quality on another.

\MODEL{} uses this evidence through a joint prior on the latent qualities of each item, whose correlation matrix links the dimensions.
Let $\theta_j=(\theta_{j1},\ldots,\theta_{jD})$ collect the latent qualities of item $j$.
The prior is a multivariate normal distribution with unit variances:
\begin{equation}
\theta_j \sim \mathcal{N}(0, R),
\qquad
R_{dd}=1 \text{ for every } d,
\label{eq:corr}
\end{equation}
where $\mathcal{N}(0,R)$ is the multivariate normal distribution with mean zero and covariance matrix $R$, and the off-diagonal entry $R_{dd'}$ is the latent correlation between dimensions $d$ and $d'$.
The unit variances fix the scale of each dimension, as the standard normal prior does in one-dimensional IRT.
If $R$ is the identity matrix, or if the benchmark has one dimension, \MODEL{} reduces to one independent graded-response model per dimension, which we write \MODEL{} ($R{=}I$).

\subsection{Estimation}
\label{sec:estim}

We now estimate the judge parameters $\alpha=(\alpha_{id})$ and $\beta=(\beta_{idk})$ and the correlation matrix $R$, collected in $\xi=(\alpha,\beta,R)$, from the ratings $Y$.
Because the latent qualities are not observed, we also compute the posterior of the latent qualities of each item.
The parameter updates average over this posterior.

\emph{Objective.}
We choose $\xi$ to maximize the marginal likelihood of the ratings, which averages over the latent qualities:
\begin{equation}
  L(\xi)=\prod_{j=1}^{N}\int \underbrace{\mathcal{N}(\theta_j;0,R)}_{\text{prior}}\underbrace{\prod_{d,i}P(Y_{ijd}\mid\theta_{jd};\alpha_{id},\beta_{id})}_{\text{likelihood}}\,d\theta_j,
  \label{eq:marglik}
\end{equation}
where $\mathcal{N}(\theta_j;0,R)$ is the density of $\mathcal{N}(0,R)$ at $\theta_j$.
The integral has no closed form, so we maximize $L$ with an expectation--maximization (EM) procedure that repeats three steps:
the E-step computes the posterior of the latent qualities under the current $\xi$, and the M-step and the correlation update then re-estimate $\xi$ from this posterior.

\paragraph{E-step.}
By Bayes' rule, the posterior $q_j$ of the latent qualities of item $j$ is proportional to the prior times the likelihood in Eq.~\eqref{eq:marglik}.
$q_j$ has no closed form, so we approximate it numerically.
In one dimension, we can evaluate it on a fixed grid $\mathcal{G}$ of latent values, but in $D$ dimensions, a joint grid needs $|\mathcal{G}|^D$ points.
We therefore approximate $q_j$ in two steps.
First, we ignore the correlation between dimensions and compute a separate posterior for each dimension on the grid:
\[
q^{(0)}_{jd}(g)\propto\phi(g)\prod_{i}P(Y_{ijd}\mid\theta_{jd}=g),\qquad g\in\mathcal{G},
\]
where $\phi$ is the standard normal density.
This posterior uses all ratings on dimension $d$.
The product of these posteriors over $d$ differs from $q_j$ only in the prior, which treats the dimensions as independent.
Second, we correct for the correlation.
We draw $S$ vectors $\theta_j^{(1)},\ldots,\theta_j^{(S)}$ and sample each coordinate $\theta_{jd}^{(s)}$ from $q^{(0)}_{jd}$, independently of the other coordinates.
These draws follow the product $\prod_d q^{(0)}_{jd}$, not $q_j$.
Importance sampling corrects this difference: each draw receives a weight proportional to $q_j$ divided by $\prod_d q^{(0)}_{jd}$ at that draw.
The two distributions contain the same rating likelihoods, so the likelihoods cancel.
The weight is therefore the correlated prior divided by the independent prior:
\begin{equation}
\omega_j^{(s)}=\frac{r_j^{(s)}}{\sum_{t=1}^{S}r_j^{(t)}},
\qquad
r_j^{(s)}=\frac{\mathcal{N}\big(\theta_j^{(s)};0,R\big)}{\prod_{d}\phi\big(\theta_{jd}^{(s)}\big)}.
\label{eq:iw}
\end{equation}
The weighted draws approximate $q_j$.
The posterior of dimension $d$ at a grid value $g$ is the total weight of the draws whose coordinate on $d$ equals $g$:
\[
q_{jd}(g)=\sum_{s:\,\theta_{jd}^{(s)}=g}\omega_j^{(s)}.
\]
When $R=I$, $q_j$ is the product of the $q^{(0)}_{jd}$, so we set $q_{jd}=q^{(0)}_{jd}$ without sampling.

\paragraph{M-step.}
Given the posteriors $q_{jd}$, the M-step estimates the discrimination $\alpha_{id}$ and the thresholds $\beta_{id}$ of each judge and dimension.
By the conditional independence of \S\ref{sec:base}, the log-likelihood of the ratings given the latent qualities is a sum of terms, one for each judge, dimension, and item.
The parameters $\alpha_{id}$ and $\beta_{id}$ appear only in the terms of judge $i$ on dimension $d$, so each pair is updated separately.
Holding $q_{jd}$ fixed, we minimize over $\alpha_{id}>0$ and ordered $\beta_{id}$:
\begin{equation}
  (\hat\alpha_{id},\hat\beta_{id})=\operatorname*{arg\,min}\Big\{\frac{\alpha_{id}^{2}}{2s_{id}^{2}}-\sum_{j,g}q_{jd}(g)\log P(Y_{ijd}\mid g)\Big\},
  \label{eq:mstep}
\end{equation}
where $P(Y_{ijd}\mid g)$ is the category probability of Eq.~\eqref{eq:cat} at $\theta_{jd}=g$, and the sum runs over the items and the grid values.
The penalty with scale $s_{id}^2>0$ keeps the discrimination of a nearly deterministic judge finite (App.~\ref{app:implementation}).

\paragraph{Correlation update.}
The correlation update estimates $R$ from the weighted draws of the E-step.
EM maximizes the expected log-likelihood, which is the joint log-likelihood of the ratings and the latent qualities averaged over the posteriors $q_j$.
Only its prior term, $\sum_j\mathbb{E}_{q_j}[\log\mathcal{N}(\theta_j;0,R)]$, depends on $R$.
This term depends on the latent qualities only through the quadratic form $\theta_j^{\top}R^{-1}\theta_j$, so it needs only their posterior second moments:
\[
M=\sum_{j=1}^{N}\mathbb{E}_{q_j}\big[\theta_j\theta_j^{\top}\big]=\sum_{j=1}^{N}\sum_{s=1}^{S}\omega_j^{(s)}\theta_j^{(s)}\theta_j^{(s)\top}.
\]
Over all covariance matrices, the prior term is maximized by $M/N$.
To stabilize this estimate on small sets, we combine it with the identity matrix $I$, which represents independent dimensions, at a fixed weight of $D+1$ pseudo-items.
We then rescale the result to unit diagonal, as the prior requires:
\begin{equation}
\begin{aligned}
  \widetilde R
  &=\frac{N}{N+D+1}\,\frac{M}{N}+\frac{D+1}{N+D+1}\,I,\\
  R_{de}
  &=\frac{\widetilde R_{de}}{\sqrt{\widetilde R_{dd}\widetilde R_{ee}}}.
\end{aligned}
\label{eq:rupdate}
\end{equation}

\subsection{Item scores}
\label{sec:readout}

We now describe how to compute the score $\hat Y_{jd}$ from the ratings and the final estimates $\hat\xi$.
The computation first estimates the latent quality by its posterior mean and then transforms this estimate to the rating scale.

\emph{Step 1: posterior mean.}
From the weighted draws of the final E-step, the estimated quality is the weighted average of their coordinates on dimension $d$:
\begin{equation}
  \hat\theta_{jd}=\sum_{s=1}^{S}\omega_j^{(s)}\theta_{jd}^{(s)}\approx\mathbb{E}_{\hat\xi}[\theta_{jd}\mid Y_j].
  \label{eq:postmean}
\end{equation}
Each draw adds its weight to the grid value it takes, so $\hat\theta_{jd}$ equals the average of the grid values weighted by the posterior $q_{jd}$.

\emph{Step 2: rating scale.}
The latent scale has no unit, so we map $\hat\theta_{jd}$ back to the original rating scale.
At quality $\hat\theta_{jd}$, the fitted judge $i$ gives the expected rating $\sum_k k\,P(Y_{ijd}=k\mid\hat\theta_{jd};\hat\alpha_{id},\hat\beta_{id})$ (Eq.~\eqref{eq:cat}), and the score averages these expected ratings over the judges:
\begin{equation}
  \hat Y_{jd}
  =
  \frac{1}{J}\sum_{i=1}^{J}
  \sum_{k=0}^{K_d-1}
  k\,P\big(Y_{ijd}=k\mid\hat\theta_{jd};\hat\alpha_{id},\hat\beta_{id}\big).
  \label{eq:tcc}
\end{equation}
This map is the test characteristic curve \citep{lord1980}.
It averages the expected ratings at $\hat\theta_{jd}$, not the observed ratings, so the judge weighting of Step~1 carries over.
All discriminations are positive, so the map is increasing, and $\hat Y_{jd}$ ranks the items in the same order as $\hat\theta_{jd}$.

\emph{How the score weights each judge.}
Eq.~\eqref{eq:postmean} does not show how much each judge contributes to the score.
To see it, let $\tilde Y_{ijd}$ be the latent value that best explains the rating of judge $i$ under its own thresholds.
Near this value, the log-likelihood of the rating is approximately quadratic with curvature $w_{ijd}$, the Fisher information of judge $i$, so the posterior mean is approximately a weighted average of the values $\tilde Y_{ijd}$:
\begin{equation}
  \hat\theta_{jd}
  \approx
  \frac{\sum_{i} w_{ijd}\,\tilde Y_{ijd}}{\sum_{i} w_{ijd}},
  \qquad
  w_{ijd}=\alpha_{id}^2\,v_{ijd}(\hat\theta_{jd}).
  \label{eq:readout}
\end{equation}
The factor $\alpha_{id}^2$ depends only on the reliability of judge $i$ and is the same for every item, so a judge that rates at random gets almost no weight, as in Figure~\ref{fig:example}.
The factor $v_{ijd}$ depends on where the item lies on the latent scale: it is large near the judge's thresholds, where its rating separates items well, and small where the judge gives almost every item the same rating.

\section{When \MODEL{} Is More Accurate}
\label{sec:theory}

To compare aggregators, we first formalize their accuracy and introduce an oracle that attains the highest possible accuracy (\S\ref{sec:ceiling}).
The gain of \MODEL{} over the mean increases with the reliability variation of the judges (\S\ref{sec:cost}), and Dawid--Skene requires several times as many items as \MODEL{} to attain the same accuracy when both models are correct (\S\ref{sec:ds}).
All proofs are in App.~\ref{app:proofs}.

\subsection{The accuracy ceiling}
\label{sec:ceiling}

We first define an \emph{oracle} that knows the true judge parameters, and we use its accuracy as a common benchmark for every aggregator.
We analyze the aggregators under the model of \S\ref{sec:method}:
\begin{assumption}[Rating model]
\label{ass:model}
The items are independent, and the latent qualities of each item satisfy $\theta_j\sim\mathcal{N}(0,R^{\ast})$ with unit diagonal.
Given $\theta_j$, the ratings of different judges and dimensions are independent, and $Y_{ijd}$ follows Eq.~\eqref{eq:cat} with discrimination $\alpha^{\ast}_{id}>0$ and thresholds $\beta^{\ast}_{id}$.
\end{assumption}

To compare aggregators, we define what an aggregator is and how we measure its accuracy.
Fix a target dimension $d$, and let $Y_j=(Y_{ijd})_{i,d}\in\mathbb{R}^{J\times D}$ collect the ratings of item $j$ by all judges on all dimensions.
An aggregator $f$ maps $Y_j$ to a scalar score $f(Y_j)$ that estimates the latent quality $\theta_{jd}$.
It may use the ratings on every dimension, as \MODEL{} does through $R$, but its output is a single score for dimension $d$.
We measure accuracy by the squared correlation between the score and the latent quality across items:
\[
\mathrm{Acc}_d(f)=\operatorname{corr}^2\big(f(Y_j),\theta_{jd}\big).
\]
The oracle scores item $j$ by its posterior mean under the true parameters $\xi^{\ast}=(\alpha^{\ast},\beta^{\ast},R^{\ast})$, $m^{\ast}(Y_j)=\mathbb{E}[\theta_{jd}\mid Y_j]$.
Among all functions of the observations, the conditional mean has the highest correlation with the variable \citep{renyi1959}, so no aggregator is more accurate than the oracle.
We call the accuracy of the oracle, $\mathrm{Acc}_d(m^{\ast})$, the \emph{accuracy ceiling}.
Every aggregator $f$ thus lies below the ceiling by the gap $\mathrm{Acc}_d(m^{\ast})-\mathrm{Acc}_d(f)\ge0$. For the mean, we write
\[
\Delta^{\mathrm{mean}}_d=\mathrm{Acc}_d(m^{\ast})-\mathrm{Acc}_d(\text{mean})\ge0.
\]

\subsection{Comparison with the mean}
\label{sec:cost}

The mean ignores how the judges differ, both in reliability and in their use of the rating scale, and \MODEL{} gains over the mean by modeling both.

Let $m_\xi(Y_j)=\mathbb{E}_\xi[\theta_{jd}\mid Y_j]$ be the posterior mean under parameters $\xi$, and let $\hat\xi_N$ be the estimate of \MODEL{} from $N$ items.
\MODEL{} must estimate $\xi$, so it gains over the mean only when the accuracy lost in this estimation is smaller than $\Delta^{\mathrm{mean}}_d$.
The next result measures this loss:
\begin{theorem}[Estimation loss]
\label{thm:cost}
Let Assumption~\ref{ass:model} hold with $\operatorname{Var}m^{\ast}(Y_j)>0$ and $\Delta^{\mathrm{mean}}_d>0$.
Let $\xi^{\ast}$ be an interior point, and let $\sqrt N(\hat\xi_N-\xi^{\ast})\xrightarrow{d}\mathcal{N}\big(0,\mathcal{I}(\xi^{\ast})^{-1}\big)$, where the Fisher information $\mathcal{I}(\xi^{\ast})$ of one item is nonsingular.
Let $H=-\nabla^2_\xi\mathrm{Acc}_d(m_\xi)$ at $\xi=\xi^{\ast}$.
\begin{enumerate}[label=(\alph*),leftmargin=*,itemsep=0pt,topsep=2pt]
  \item $N\big(\mathrm{Acc}_d(m^{\ast})-\mathrm{Acc}_d(m_{\hat\xi_N})\big)\xrightarrow{d}L$ with $\mathbb{E}[L]=\kappa=\tfrac12\operatorname{tr}\big(H\mathcal{I}(\xi^{\ast})^{-1}\big)$.
  \item $P\big(\mathrm{Acc}_d(m_{\hat\xi_N})>\mathrm{Acc}_d(\mathrm{mean})\big)\to1$.
\end{enumerate}
\end{theorem}
Part (a) says that \MODEL{} loses about $\kappa/N$ of accuracy on $N$ items, whereas the mean loses $\Delta^{\mathrm{mean}}_d$ for every $N$.
Comparing the two losses gives the \emph{break-even size} $N_c$:
\begin{equation}
\frac{\kappa}{N}<\Delta^{\mathrm{mean}}_d
\iff
N>N_c=\frac{\kappa}{\Delta^{\mathrm{mean}}_d}.
\label{eq:nc}
\end{equation}
Above $N_c$, \MODEL{} is more accurate than the mean. 
Part (b) shows that it is more accurate with probability tending to one as $N$ grows.
The break-even size is small exactly when the gap $\Delta^{\mathrm{mean}}_d$ is large, so \MODEL{} overtakes the mean with few items wherever the mean loses much accuracy (\S\ref{sec:main}).

The gap $\Delta^{\mathrm{mean}}_d$ has no closed form under the graded-response model, so we study it in the one-factor model of classical test theory \citep{lordnovick1968} (App.~\ref{app:onefactor}), where the reliability of a judge is the correlation between its ratings and the latent quality.
Let $\bar r$ be the mean reliability of the judges, and define the \emph{reliability variation} $\delta_r$ as the standard deviation of the reliabilities divided by $\bar r$.
The next result shows that as the judges differ more in reliability, the gap of the mean grows and the break-even size falls:
\begin{proposition}[Gap of the mean]
\label{prop:nc}
In the one-factor model with $J\ge3$ judges, as $\delta_r\to0$,
\[
\Delta^{\mathrm{mean}}_d=\tau\,\delta_r^2+O(\delta_r^3),\qquad N_c=\frac{\kappa_0}{\tau\,\delta_r^2}\big(1+O(\delta_r)\big),
\]
where $\tau>0$ and the value $\kappa_0>0$ of $\kappa$ at equal reliabilities depend only on $J$ and $\bar r$.
\end{proposition}
The gap is zero when all judges are equally reliable, and the number of items that \MODEL{} needs falls as $\delta_r^{-2}$.
Since $N_c$ scales as $\delta_r^{-2}$, the ratio $N/N_c$ is proportional to $N\delta_r^2$, and the next result shows that this quantity alone determines the probability that \MODEL{} is more accurate than the mean:
\begin{theorem}[Winning probability]
\label{thm:region}
In the one-factor model with $J\ge3$ judges, let $\hat\xi_N$ be the maximum-likelihood estimate of the one-factor model, and let the reliabilities deviate from their mean $\bar r$ by fixed proportions scaled by $N^{-1/2}$, so that $N\delta_r^2$ stays fixed (App.~\ref{app:onefactor}). Then
\[
P\big(\mathrm{Acc}_d(m_{\hat\xi_N})>\mathrm{Acc}_d(\mathrm{mean})\big)\to F_{J-1}\big(\tfrac{(J-1)\tau}{\kappa_0}N\delta_r^2\big),
\]
where $F_{J-1}$ is the chi-square distribution function with $J-1$ degrees of freedom.
\end{theorem}
Thus, the more the judges differ in reliability, the fewer items \MODEL{} needs to be more accurate than the mean with high probability.

\subsection{Comparison with Dawid--Skene}
\label{sec:ds}

We now compare \MODEL{} with Dawid--Skene.
Dawid--Skene treats the latent quality as one of $K$ values, one per rating category, so we consider two cases: a continuous quality and a discrete quality with $K$ values.
Since Dawid--Skene models each dimension separately, we consider one dimension and drop the index $d$.

If the latent quality is continuous, as under Assumption~\ref{ass:model}, the model of Dawid--Skene is wrong, and we show that its accuracy stays below the ceiling for every $N$, whereas that of \MODEL{} approaches it.
The estimate of Dawid--Skene converges to the parameter of its model closest to the true distribution of the ratings \citep{white1982}, and we write $\Delta^{\mathrm{DS}}\ge0$ for the gap of the resulting score below the ceiling:
\begin{proposition}[Floor of Dawid--Skene]
\label{prop:dsmain}
Let Assumption~\ref{ass:model} hold, and let the estimates of \MODEL{} and Dawid--Skene be asymptotically normal around their limits.
Also, let $\hat f^{\mathrm{DS}}_N$ be the aggregator that Dawid--Skene fits to $N$ items.
Then, as $N\to\infty$, $\mathrm{Acc}(m^{\ast})-\mathrm{Acc}(\hat f^{\mathrm{DS}}_N)\to\Delta^{\mathrm{DS}}$ in probability, and if $\Delta^{\mathrm{DS}}>0$,
\[
P\big(\mathrm{Acc}(m_{\hat\xi_N})>\mathrm{Acc}(\hat f^{\mathrm{DS}}_N)\big)\to1.
\]
\end{proposition}

If the latent quality takes one of $K$ values, both models can be correct, and we compare the two methods in this case:
\begin{assumption}[$K$-value quality]
\label{ass:klevel}
The latent quality $\theta_j$ takes one of $K$ equally spaced values with probabilities $\pi^{\ast}_c>0$.
Given $\theta_j$, the ratings of the $J\ge3$ judges are independent, and $Y_{ij}$ follows Eq.~\eqref{eq:cat} with $\alpha^{\ast}_i>0$ and ordered thresholds $\beta^{\ast}_i$.
\end{assumption}
Let \MODEL{}-$K$ denote \MODEL{} with its normal prior replaced by these $K$ values with free probabilities, and let $\hat f^{K}_N$ be the aggregator that it fits to $N$ items.
The two methods then differ in the cost of estimating their parameters.
\begin{theorem}[Efficiency comparison]
\label{thm:ds}
Let Assumption~\ref{ass:klevel} hold, and let both models be identifiable at the true parameter with nonsingular Fisher information, so that their maximum-likelihood estimates are asymptotically normal \citep{vandervaart1998}.
\begin{enumerate}[label=(\alph*),leftmargin=*,itemsep=0pt,topsep=2pt]
  \item The losses below the ceiling converge in distribution,
  \[
  \begin{aligned}
  N\big(\mathrm{Acc}(m^{\ast})-\mathrm{Acc}(\hat f^{K}_N)\big)&\xrightarrow{d}L_K,\\
  N\big(\mathrm{Acc}(m^{\ast})-\mathrm{Acc}(\hat f^{\mathrm{DS}}_N)\big)&\xrightarrow{d}L_{\mathrm{DS}},
  \end{aligned}
  \]
  and the means $\kappa_K=\mathbb{E}[L_K]$ and $\kappa_{\mathrm{DS}}=\mathbb{E}[L_{\mathrm{DS}}]$ satisfy $\kappa_K\le\kappa_{\mathrm{DS}}$.
  \item If $K=2$, then $\kappa_{\mathrm{DS}}=\kappa_K$. If $3\le K\le10$, then $\kappa_{\mathrm{DS}}>\kappa_K$ at almost every parameter.
\end{enumerate}
\end{theorem}
Dawid--Skene therefore needs $\kappa_{\mathrm{DS}}/\kappa_K$ times as many items as \MODEL{}-$K$ for the same accuracy.
This is because Dawid--Skene does not use the order of the rating scale, so it estimates $K(K-1)$ rating probabilities per judge, whereas \MODEL{}-$K$ estimates $K$ parameters per judge.

\section{Evaluation}
\label{sec:eval}

We evaluate \MODEL{} on real and simulated datasets.

\subsection{Setup}
\label{sec:setup}

\paragraph{Datasets.}
We evaluate on three meta-evaluation benchmarks.
\textbf{SummEval} \citep{fabbri2021summeval} contains $1{,}600$ news summaries rated by three experts on four dimensions (coherence, consistency, fluency, relevance) on a 1--5 scale.
\textbf{Topical-Chat} \citep{mehri2020usr} contains $360$ dialogue responses rated by three annotators on six dimensions with 2-, 3-, and 5-point scales.
\textbf{WMT22} en--de \citep{freitag2022wmt22} contains $19{,}725$ translated segments with professional quality scores on one dimension.
To test the small-$N$ regime, we draw $15$ random subsets of $N=50$ items, and we also evaluate on all items.

\paragraph{Judge pools.}
Each judge is a large language model queried with one fixed prompt.
The candidate set consists of ten models from proprietary and open-weight families (App.~\ref{app:prompts}).
We draw $100$ random pools of five judges from the candidate set with a fixed seed.

\paragraph{Models.}
We compare \MODEL{} with seven baseline aggregators.
Three of them estimate no parameters: the mean, the median, and the trimmed mean, which discards the highest and the lowest $20\%$ of the ratings of each item.
The other four estimate parameters from the ratings: Dawid--Skene \citep{dawid1979}, MACE \citep{hovy2013mace}, GLAD \citep{whitehill2009glad}, and CARE \citep{zhao2026care}.

\paragraph{Metrics.}
We report the Spearman correlation $\rho$ between the scores of each method and the human ratings, averaged over dimensions and then over pools.

\subsection{Main results}
\label{sec:main}

\begin{table}[t]
  \centering\footnotesize
  \setlength{\tabcolsep}{2.2pt}
  \begin{tabular}{l cc cc cc}
  \toprule
  & \multicolumn{2}{c}{\textbf{SummEval}} & \multicolumn{2}{c}{\textbf{Topical-Chat}} & \multicolumn{2}{c}{\textbf{WMT22}} \\
  \cmidrule(lr){2-3}\cmidrule(lr){4-5}\cmidrule(lr){6-7}
  Method & small & full & small & full & small & full \\
  \midrule
  Mean                  & .518 & .543 & .715 & .703 & .441 & .434\\
  Median                & .524 & .548 & .683 & .669 & .430 & .427\\
  Trimmed Mean          & .536 & .558 & .718 & .705 & .439 & .438\\
  Dawid--Skene          & .514 & .561 & .702 & .717 & .420 & .438\\
  CARE                  & .518 & .559 & .701 & .709 & .436 & .437\\
  GLAD                  & .510 & .537 & .718 & .712 & .445 & .437\\
  MACE                  & .528 & .540 & .718 & .712 & .443 & .434\\
  \MODEL{}              & \textbf{.545} & \textbf{.568} & \textbf{.728} & \textbf{.729} & \textbf{.450} & \textbf{.439}\\
  \bottomrule
  \end{tabular}
  \caption{Accuracy of aggregators on random judge pools: Spearman $\rho$ against the human ratings, averaged over dimensions and pools. \emph{small}: $N{=}50$, mean over $15$ random subsets. \emph{full}: all items. \textbf{Bold}: highest value in each column.}
  \label{tab:main}
\end{table}

Table~\ref{tab:main} reports the $\rho$ of each aggregator, and \MODEL{} has the highest $\rho$ in every column.
We compare two methods with a paired Wilcoxon signed-rank test over the $100$ pools.
Against the mean, \MODEL{} gains $+0.025$ to $+0.027$ on SummEval and $+0.013$ to $+0.025$ on Topical-Chat, and is higher in $96$--$100\%$ of pools at both sizes ($p<10^{-16}$).
On WMT22, it is higher in $95\%$ of pools at $N{=}50$ and in every pool at full $N$ ($p<10^{-16}$).
Against Dawid--Skene, \MODEL{} gains $+0.030$, $+0.026$, and $+0.030$ at $N{=}50$ on the three benchmarks.

\begin{figure}[t]
  \centering
  \includegraphics[width=\columnwidth]{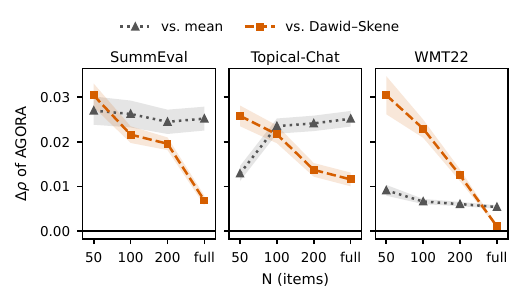}
  \caption{Accuracy gain of \MODEL{} over the mean and Dawid--Skene as the evaluation set grows (with $95\%$ confidence bands).}
  \label{fig:ncurve}
\end{figure}

\paragraph{Gain against evaluation-set size.}
\label{sec:efficiency}
We repeat the comparison at $N\in\{50,100,200\}$ and at full $N$ on the same pools (Figure~\ref{fig:ncurve}).
\MODEL{} exceeds the mean at every size.
Its margin over Dawid--Skene shrinks as $N$ grows but stays positive at full $N$, in line with \S\ref{sec:ds}.

\paragraph{Judges from the same family.}
\MODEL{} assumes that the ratings of different judges are independent given the latent quality (Assumption~\ref{ass:model}).
Judges from the same model family may share errors and so violate this assumption.
In the $39$--$46$ pools per benchmark that contain such a pair (\texttt{gpt-5} and \texttt{gpt-oss-120b}, or \texttt{gemini-2.5-pro} and \texttt{gemma-4-31B}), \MODEL{} exceeds the mean in $96\%$ of pools at both sizes on average.

\subsection{Verifying the theory}
\label{sec:synthetic}

This subsection verifies the two statements of \S\ref{sec:cost}.
First, \MODEL{} approaches the ceiling as the set grows, whereas the mean stays a fixed distance below it (Thm.~\ref{thm:cost}).
Second, \MODEL{} needs fewer items to beat the mean when the judges differ more in reliability (Thm.~\ref{thm:region}).
The human ratings do not reveal the accuracy ceiling, so we use simulated ratings with a known ceiling.

\begin{figure}[t]
  \centering
  \includegraphics[width=\columnwidth]{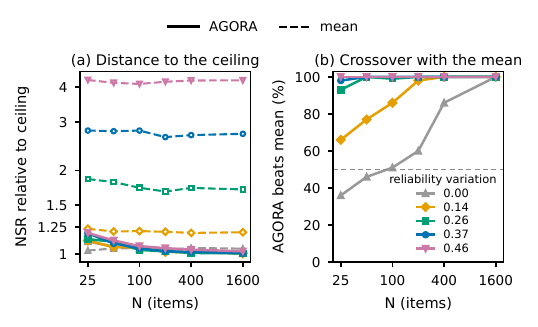}
  \caption{Synthetic ratings from a known model, $100$ draws per point; colors and markers mark the reliability variation. (a) Median NSR of \MODEL{} (solid) and of the mean (dashed), divided by that of the oracle. (b) Percentage of draws in which \MODEL{} is more accurate than the mean.}
  \label{fig:synth}
\end{figure}

We draw ratings from a known graded-response model with five judges, at five levels of reliability variation and at set sizes from $25$ to $1{,}600$ items (App.~\ref{app:synth}).
Because the true parameters are known, we can compute the ceiling and compare \MODEL{} and the mean with it.
We measure the distance to the ceiling by the noise-to-signal ratio $\NSR(f)=(1-\mathrm{Acc}_d(f))/\mathrm{Acc}_d(f)$ of the score $f(Y_j)$ of an aggregator, which is zero when the score is a linear function of the latent quality and grows as the score becomes noisier.
Figure~\ref{fig:synth}a divides it by the NSR of the oracle, so a value of one means the accuracy of the oracle.

Figure~\ref{fig:synth}a confirms the first statement.
The NSR of \MODEL{} exceeds that of the oracle by $11$--$19\%$ at $N=25$ and by $2\%$ at full size.
The ratio of the mean stays far above one at every set size, and this excess grows with the reliability variation.
Figure~\ref{fig:synth}b confirms the second statement: the winning probability of Thm.~\ref{thm:region} rises faster with $N$ as the reliability variation grows.
At a variation of $0.26$ or more, \MODEL{} wins in at least $93\%$ of the draws already at $25$ items.

\subsection{Ablation study}
\label{sec:ablation}

\paragraph{Weighting methods.}
We replace the discriminations of \MODEL{} with five weighting heuristics that need no latent-trait model: \emph{Most-agreeing judge}, the single judge that agrees most with the others; \emph{Drop least-agreeing}, the mean without the judge that agrees least; \emph{Agreement-weighted}, a weighted mean of standardized ratings with weights equal to each judge's agreement with the others; \emph{Inverse-variance}, an inverse-variance weighting by each judge's disagreement with the others; and \emph{Spectral} \citep{parisi2014spectral}, the weighting by the leading eigenvector of the judge correlation matrix, a principal-component approximation of one-factor weights.

\begin{table}[t]
  \centering\footnotesize
  \setlength{\tabcolsep}{2.2pt}
  \begin{tabular}{l cc cc cc}
  \toprule
  & \multicolumn{2}{c}{\textbf{SummEval}} & \multicolumn{2}{c}{\textbf{Topical-Chat}} & \multicolumn{2}{c}{\textbf{WMT22}} \\
  \cmidrule(lr){2-3}\cmidrule(lr){4-5}\cmidrule(lr){6-7}
  Method & small & full & small & full & small & full \\
  \midrule
  Most-agreeing judge   & .530 & .560 & .674 & .672 & .440 & .415\\
  Drop least-agreeing   & .538 & .562 & .719 & .710 & \textbf{.452} & .437\\
  Agreement-weighted    & .534 & .556 & .718 & .709 & .448 & .436\\
  Inverse-variance      & .537 & .559 & .721 & .713 & .451 & .437\\
  Spectral              & .534 & .556 & .717 & .708 & .446 & .435\\
  \midrule
  Equal discrimination  & .527 & .551 & .713 & .703 & .445 & .435\\
  \MODEL{} ($R{=}I$)    & .540 & .564 & .723 & .718 & .450 & \textbf{.439}\\
  \MODEL{}              & \textbf{.545} & \textbf{.568} & \textbf{.728} & \textbf{.729} & .450 & \textbf{.439}\\
  \bottomrule
  \end{tabular}
  \caption{Ablation study on the pools of Table~\ref{tab:main} (Spearman $\rho$ with the human ratings). Top: weighting methods. Bottom: thresholds only, then discriminations, then the correlation across dimensions. \textbf{Bold}: best in each column.}
  \label{tab:ablation}
\end{table}

On SummEval and Topical-Chat, \MODEL{} exceeds the best heuristic by $0.006$ to $0.016$ in $\rho$ at both sizes (Table~\ref{tab:ablation}).
On WMT22, it exceeds every heuristic at full $N$ and is within $0.002$ of the two strongest heuristics on small sets, \emph{Drop least-agreeing} and \emph{Inverse-variance}.

\paragraph{Model components.}
\MODEL{} has three components: the thresholds $\beta_{id}$, which model the leniency of each judge and its use of the rating scale; the discriminations $\alpha_{id}$, which weight each judge by its reliability; and the correlation $R$ across dimensions.
We add the components one at a time and evaluate each version on the pools of Table~\ref{tab:main}.
\emph{Equal discrimination} keeps only the thresholds: it fits them but gives every judge the same discrimination in the item scores.
\MODEL{} ($R{=}I$) adds the discriminations, and \MODEL{} adds the correlation across dimensions.

The discriminations give the largest gain: adding them raises $\rho$ by $0.004$ to $0.015$ (Table~\ref{tab:ablation}).
Adding the correlation across dimensions raises $\rho$ by another $0.004$ to $0.011$ on SummEval and Topical-Chat.

\subsection{Extension to pairwise comparison}
\label{sec:pairwise}

We extend \MODEL{} from pointwise ratings to pairwise verdicts, since many LLM-judge protocols ask for pairwise comparisons.
The reliability weighting of \MODEL{} also applies to this format (App.~\ref{app:pairwise}).
We replace the graded-response likelihood by a Bradley--Terry \citep{bradley1952rank} likelihood of each verdict, scaled by a per-judge discrimination.
An optional tie threshold models tied verdicts.
A label-free criterion, the held-out log-likelihood on $20\%$ of the verdicts, decides whether to use the tie threshold and the discrimination penalty of Eq.~\eqref{eq:mstep}.
We derive the verdicts from the ratings of the same $100$ pools: a judge prefers the item it rates higher and ties items it rates equally.
We compare with Bradley--Terry and with BT-$\gamma$, the Bradley--Terry model with a per-judge discrimination \citep{btsigma2026,xu2026judgeaware}.

\begin{table}[t]
  \centering\footnotesize
  \setlength{\tabcolsep}{2.6pt}
  \begin{tabular}{l cc cc cc}
  \toprule
  & \multicolumn{2}{c}{\textbf{SummEval}} & \multicolumn{2}{c}{\textbf{Topical-Chat}} & \multicolumn{2}{c}{\textbf{WMT22}} \\
  \cmidrule(lr){2-3}\cmidrule(lr){4-5}\cmidrule(lr){6-7}
  Method & small & full & small & full & small & full \\
  \midrule
  Bradley--Terry   & .521 & .543 & .708 & .685 & .444 & .359\\
  BT-$\gamma$      & .536 & .557 & .719 & .700 & .445 & .349\\
  Pairwise \MODEL{} & \textbf{.538} & \textbf{.558} & \textbf{.721} & \textbf{.705} & \textbf{.449} & \textbf{.363}\\
  \bottomrule
  \end{tabular}
  \caption{Pairwise verdicts from the ratings of the pools of Table~\ref{tab:main}: Spearman $\rho$ against the human ratings at $N=50$ and full $N$.}
  \label{tab:pairwise}
\end{table}

Pairwise \MODEL{} matches or exceeds BT-$\gamma$ in every column of Table~\ref{tab:pairwise}, by $0.001$ to $0.014$ in $\rho$, so \MODEL{} extends to pairwise verdicts without loss of accuracy.

\section{Conclusion}
\MODEL{} aggregates LLM judges of uneven reliability with a graded-response model whose discriminations act as reliability weights and whose latent correlation matrix shares evidence across dimensions.
Its gain over the mean grows with the reliability variation of the judges, and the number of items it needs to overtake the mean falls accordingly.
Even when both models are correct, Dawid--Skene needs several times as many items as \MODEL{} to reach the same accuracy.
On 100 random judge pools over three benchmarks, \MODEL{} has the highest Spearman correlation with human ratings among standard aggregators at small and full sizes, and synthetic data confirm the theory.

\section*{Impact Statement}
This work makes evaluation with LLM judges more reliable when few human labels are available, which can lower the cost of evaluating generative systems.
Because the aggregate is learned from the judges alone, it inherits biases that all judges share, so it should complement rather than replace human evaluation in high-stakes settings.
We see no other societal consequences that must be specifically highlighted here.

\bibliographystyle{icml2027}
\bibliography{references}

\newpage
\appendix
\onecolumn
\renewcommand{\rev}[1]{#1}
\renewcommand{\revred}[1]{#1}
\renewcommand{\revp}[1]{#1}
\renewcommand{\pol}[1]{#1}
\renewcommand{\opt}[1]{#1}

\section{Estimation Details}
\label{app:estimation}
\rev{\MODEL{} estimates the judge parameters by marginal maximum likelihood and the correlation matrix by Eq.~\eqref{eq:rupdate}, with the EM procedure of Algorithm~\ref{alg:agora}.}

\subsection{\rev{Fitting and scoring procedure}}
\label{app:algorithm}
\rev{Algorithm~\ref{alg:agora} lists the steps of \MODEL{}: the initialization, the EM iterations of \S\ref{sec:estim}, and the item scores of \S\ref{sec:readout}.}

\nolinenumbers
\begin{algorithm}[h]
\small
\caption{\rev{Fitting and scoring with \MODEL{}}}
\label{alg:agora}
\begin{algorithmic}[1]
\REQUIRE ratings $Y_{ijd}$ for $N$ items, $J$ judges, $D$ dimensions\rev{; number of iterations $T$}
\ENSURE per-item scores $\hat Y_{jd}$
\STATE $\alpha\gets 1$; thresholds $\beta$ from one \rev{graded-response} step on the mean; $R\gets I$
\STATE \textbf{E-step:} posterior $q_{jd}$ at the initial parameters
\FOR{$t=1$ \textbf{to} $T$}
  \STATE \textbf{M-step:} per judge and dimension, minimize Eq.~\eqref{eq:mstep} on the expected counts
  \STATE \textbf{E-step:} grid posterior per dimension, reweighted by Eq.~\eqref{eq:iw}
  \STATE \textbf{Correlation update:} $R$ from Eq.~\eqref{eq:rupdate}
\ENDFOR
\STATE $\hat\theta_{jd}\gets$ posterior mean at the final parameters
\STATE $\hat Y_{jd}\gets$ \rev{test characteristic curve}, Eq.~\eqref{eq:tcc}
\RETURN $\hat Y$ (with $\alpha$, $\beta$, $R$)
\end{algorithmic}
\end{algorithm}
\linenumbers

\subsection{\rev{Implementation settings}}
\label{app:implementation}
The M-step minimizes Eq.~\eqref{eq:mstep} by L-BFGS-B over $\alpha_{id}\in[0.01,20]$ and the ordered thresholds, which are parameterized through cumulative positive increments.
The E-step uses a grid $\mathcal{G}$ of $150$ points on $[-4.5,4.5]$ for each dimension and $S=300$ importance samples per item for the correlated prior.
The procedure starts from $\alpha_{id}=1$ and thresholds from one graded-response step, and it runs $T=12$ iterations.
\rev{On the grid $\mathcal{G}$, the M-step objective depends on the ratings only through the expected counts \revred{$n_{idk}(g)=\sum_{j:\,Y_{ijd}=k}q_{jd}(g)$}, so its cost does not grow with $N$; with a correlated prior, $q_{jd}$ is the weighted sample of Eq.~\eqref{eq:iw}.}
\revp{The M-step adds the penalty $\alpha_{id}^2/(2s_{id}^2)$ of a regularized-horseshoe prior on the discriminations of each dimension \citep{piironen2017regularized}, with slab scale $c=5$.
The scales $s_{id}^2$ are updated in closed form once per iteration.
The penalty keeps the discrimination of a nearly deterministic judge finite.
Its strength does not grow with $N$, so it leaves the large-$N$ limit of Assumption~\ref{ass:est} unchanged.}
These settings are fixed in advance and are the same for every benchmark.
As a numerical safeguard, the implementation flips the sign of $\hat\theta_{\cdot d}$ if it correlates negatively with the mean.

\paragraph{Information factorization.}
The \rev{judge} weight in Eq.~\eqref{eq:readout} is the graded-response Fisher information
\[
I_{ijd}(\theta)=\sum_k\frac{[P'_{ijdk}(\theta)]^2}{P_{ijdk}(\theta)}=\alpha_{id}^2\,v_{ijd}(\theta),
\]
where $P_{ijdk}(\theta)=P(Y_{ijd}=k\mid\theta_{jd}=\theta)$ is the category probability of Eq.~\eqref{eq:cat} and $P'_{ijdk}$ is its derivative in $\theta$.

\subsection{\revp{Pairwise extension}}
\label{app:pairwise}
\rev{\emph{From ratings to verdicts.}
A verdict of judge $i$ on items $a$ and $b$ is an ordinal rating with three categories, $b$ preferred, tie, and $a$ preferred, coded $0$, $1$, and $2$.
Pairwise \MODEL{} applies the graded-response model of Eq.~\eqref{eq:cat} to this rating, with the score difference $s_a-s_b$ in place of the latent quality, the discrimination $\gamma_i$ in place of $\alpha_{id}$, and the thresholds $-w_i/\gamma_i$ and $w_i/\gamma_i$:
\[
P(\text{verdict}\ge1)=\sigma(z+w_i),\qquad P(\text{verdict}=2)=\sigma(z-w_i),\qquad z=\gamma_i(s_a-s_b).
\]
The probability that $b$ is preferred is thus $1-\sigma(z+w_i)=\sigma(-z-w_i)$.
With $w_i=0$ and ties dropped, the model reduces to the Bradley--Terry likelihood $\sigma(z)$ scaled by $\gamma_i$.
The weighting argument of \S\ref{sec:readout} carries over: $\gamma_i^2$ multiplies the Fisher information of every verdict of judge $i$.}

\rev{\emph{Configurations.}}
\revp{Judge $i$ compares items $a$ and $b$ through $z=\gamma_i(s_a-s_b)$, where $s_a$ is the score of item $a$ and $\gamma_i>0$ is the discrimination of the judge.
Without ties, the judge prefers $a$ with probability $\sigma(z)$, and tied verdicts are dropped.
With standard normal priors on $s_a$ and $\log\gamma_i$, this model is BT-$\gamma$.
The tie option adds a width $w_i>0$ per judge: the judge prefers $a$ with probability $\sigma(z-w_i)$, prefers $b$ with probability $\sigma(-z-w_i)$, and ties otherwise, with a standard normal prior on $\log w_i$.
The prior option replaces the prior on $\gamma_i$ by the regularized-horseshoe prior that \MODEL{} places on its discriminations.
The four configurations share the objective, the optimizer, and the starting point, so BT-$\gamma$ is the configuration with both options off.
We hold out $20\%$ of the verdicts, fit each configuration on the rest, and keep the one with the highest mean log probability of the held-out decisive verdicts.
We then refit this configuration on all verdicts.
A judge compares every pair of items when a set has at most $25{,}000$ pairs; otherwise we draw $25{,}000$ pairs at random, the same pairs for every judge.}

\section{Proofs}
\label{app:proofs}
Throughout the proofs, we write $G(\xi)=\mathrm{Acc}_d(m_\xi)$ for the accuracy of the posterior mean under parameters $\xi$, so that $G(\xi^{\ast})=\mathrm{Acc}_d(m^{\ast})$ and $H=-\nabla^2G(\xi^{\ast})$.

\subsection{The accuracy ceiling and the mean}

\paragraph{Cost of estimation.}
We call an aggregator \emph{fixed} if it uses no parameters estimated from the data, as the mean does.
We assume the standard asymptotic normality of the maximum-likelihood estimate, which holds when the model is identifiable and its likelihood is smooth \citep{vandervaart1998}:
\begin{assumption}[Regular estimation]
\label{ass:est}
The true parameter $\xi^{\ast}$ is an interior point of the parameter set, the Fisher information $\mathcal{I}(\xi^{\ast})$ of the ratings of one item is nonsingular, and
$\sqrt N(\hat\xi_N-\xi^{\ast})\to\mathcal{N}\big(0,\mathcal{I}(\xi^{\ast})^{-1}\big)$
in distribution.
\end{assumption}
Theorem~\ref{thm:cost} holds under Assumptions~\ref{ass:model} and~\ref{ass:est}. We prove part (b) in a more general form: every fixed aggregator $f$ with $\mathrm{Acc}_d(f)<G(\xi^{\ast})$ satisfies $P\big(G(\hat\xi_N)>\mathrm{Acc}_d(f)\big)\to1$. The break-even statement follows from part (a).

\paragraph{\rev{Formal statements.}}
\rev{The analysis of \S\ref{sec:cost} uses the following results, under Assumptions~\ref{ass:model} and~\ref{ass:est}.}

\begin{proposition}[Accuracy ceiling]
\label{prop:ceiling}
Under Assumption~\ref{ass:model}, every aggregator $f$ with $\operatorname{Var}f(Y_j)>0$ satisfies
\[
\mathrm{Acc}_d(f)\le\mathrm{Acc}_d(m^{\ast})=\operatorname{Var}m^{\ast}(Y_j).
\]
Equality holds if and only if $f(Y_j)=a+b\,m^{\ast}(Y_j)$ with probability one, for constants $a$ and $b\neq0$.
\end{proposition}

\begin{lemma}[\rev{Maximum of the accuracy}]
\label{lem:gmax}
\rev{Under Assumptions~\ref{ass:model} and~\ref{ass:est}, suppose $\operatorname{Var}m^{\ast}(Y_j)>0$.
The function $G$ is twice continuously differentiable near $\xi^{\ast}$, attains its maximum at $\xi^{\ast}$, and has zero gradient there, so $H=-\nabla^2G(\xi^{\ast})$ is positive semidefinite.}
\end{lemma}

\begin{corollary}[Label-free \rev{break-even} size]
\label{cor:nc}
\rev{Under Assumptions~\ref{ass:model} and~\ref{ass:est}, suppose $\operatorname{Var}m^{\ast}(Y_j)>0$ and $\Delta^{\mathrm{mean}}_d>0$.}
Let $\kappa(\xi)$ and $\Delta^{\mathrm{mean}}_d(\xi)$ denote the two quantities computed with $\xi$ as the true parameter.
Then the plug-in \rev{break-even} size $\hat N_c=\kappa(\hat\xi_N)/\Delta^{\mathrm{mean}}_d(\hat\xi_N)$ converges in probability to $N_c$.
\end{corollary}

\paragraph{Proof of Proposition~\ref{prop:ceiling}.}
Write $m=m^{\ast}(Y_j)$ and $\theta=\theta_{jd}$.
The tower property gives $\mathbb{E}[f(Y_j)\theta]=\mathbb{E}[f(Y_j)m]$ and $\mathbb{E}\theta=\mathbb{E}m$, so
\[
\operatorname{Cov}\big(f(Y_j),\theta\big)=\operatorname{Cov}\big(f(Y_j),m\big).
\]
The Cauchy--Schwarz inequality gives
\[
\operatorname{Cov}\big(f(Y_j),m\big)^2\le\operatorname{Var}f(Y_j)\,\operatorname{Var}m,
\]
with equality if and only if $f(Y_j)$ is an affine function of $m$ with probability one.
Since $\operatorname{Var}\theta=R^{\ast}_{dd}=1$, dividing by $\operatorname{Var}f(Y_j)$ gives $\mathrm{Acc}_d(f)\le\operatorname{Var}m$.
For $f=m$ the covariance equals $\operatorname{Var}m$, so $\mathrm{Acc}_d(m)=\operatorname{Var}m$.
The proof uses only the joint distribution of $(\theta_j,Y_j)$ and finite second moments.

\paragraph{Proof of the \rev{identity for human ratings} in \revp{App.~\ref{app:synth}}.}
Let the \rev{human rating} be $h_j=\nu\theta_{jd}+e_j$ with $\nu>0$ and $e_j$ independent of $(\theta_j,Y_j)$.
Then $\operatorname{Cov}(f(Y_j),h_j)=\nu\operatorname{Cov}(f(Y_j),\theta_{jd})$.
Dividing by the standard deviations gives
\[
\operatorname{corr}(f,h)=\operatorname{corr}(f,\theta_{jd})\,\frac{\nu}{\operatorname{sd}(h)},
\]
and $\nu/\operatorname{sd}(h)=\operatorname{corr}(\theta_{jd},h)$ because $\operatorname{Var}\theta_{jd}=1$.

\begin{lemma}[Uninformative judges]
\label{lem:noise}
Under Assumption~\ref{ass:model}, add judges whose ratings are independent of $(\theta_j,Y_j)$ and have positive variance.
The ceiling $\mathrm{Acc}_d(m^{\ast})$ does not change, and the accuracy of the mean strictly decreases.
\end{lemma}

\begin{proof}
Let $U_j$ collect the added ratings.
Since $U_j$ is independent of $(\theta_j,Y_j)$, the posterior mean given $(Y_j,U_j)$ equals $\mathbb{E}[\theta_{jd}\mid Y_j]$, so the ceiling does not change.
Let $S_j=\sum_iY_{ijd}$ and $T_j=\sum_kU_{kj}$, the sum of the added ratings on dimension $d$, with $\operatorname{Var}T_j>0$.
The mean of all judges is proportional to $S_j+T_j$, and $T_j$ is independent of $(\theta_j,S_j)$.
Hence
\[
\operatorname{Cov}(S_j+T_j,\theta_{jd})=\operatorname{Cov}(S_j,\theta_{jd}),\qquad\operatorname{Var}(S_j+T_j)=\operatorname{Var}S_j+\operatorname{Var}T_j.
\]
Under Assumption~\ref{ass:model}, each expected rating $\mathbb{E}[Y_{ijd}\mid\theta_{jd}]$ is strictly increasing in $\theta_{jd}$ because $\alpha^{\ast}_{id}>0$, so $\operatorname{Cov}(S_j,\theta_{jd})>0$.
The squared correlation of the mean with $\theta_{jd}$ therefore satisfies
\[
\frac{\operatorname{Cov}(S_j,\theta_{jd})^2}{\operatorname{Var}S_j+\operatorname{Var}T_j}<\frac{\operatorname{Cov}(S_j,\theta_{jd})^2}{\operatorname{Var}S_j},
\]
where the right side is its value without the added judges.
\end{proof}

\paragraph{Proof of \rev{Lemma~\ref{lem:gmax} and} Theorem~\ref{thm:cost}.}
\emph{Smoothness.}
The ratings $Y_j$ take values in a finite set $\mathcal{Y}$.
For each $y\in\mathcal{Y}$ and each interior $\xi$, the probability
\[
p_\xi(y)=\int\prod_{i,d}P_\xi(y_{id}\mid\theta)\,\phi_R(\theta)\,d\theta
\]
is positive, where $\phi_R$ is the $\mathcal{N}(0,R)$ density.
The posterior mean is
\[
m_\xi(y)=\frac{1}{p_\xi(y)}\int\theta_d\prod_{i,d}P_\xi(y_{id}\mid\theta)\,\phi_R(\theta)\,d\theta.
\]
The integrands are infinitely differentiable in $\xi$, and their derivatives are bounded by polynomials in $\theta$ times $\phi_R(\theta)$ uniformly on a neighborhood of $\xi^{\ast}$, so differentiation under the integral sign shows that $m_\xi(y)$ is twice continuously differentiable in $\xi$.
By the tower property, $\operatorname{Cov}(m_\xi(Y_j),\theta_{jd})=\operatorname{Cov}(m_\xi(Y_j),m^{\ast}(Y_j))$, where both moments are sums over $\mathcal{Y}$ weighted by $p_{\xi^{\ast}}$.
Hence
\[
G(\xi)=\frac{\operatorname{Cov}(m_\xi,m^{\ast})^2}{\operatorname{Var}m_\xi}
\]
is a twice continuously differentiable function of the finite vector $(m_\xi(y))_{y\in\mathcal{Y}}$ wherever $\operatorname{Var}m_\xi>0$.
This variance is positive at $\xi^{\ast}$ by assumption, and therefore on a neighborhood of $\xi^{\ast}$ by continuity.

\emph{\rev{Lemma~\ref{lem:gmax}}.}
Each $m_\xi$ is an aggregator, so Proposition~\ref{prop:ceiling} gives $G(\xi)\le G(\xi^{\ast})$ for every $\xi$.
Since $\xi^{\ast}$ is an interior maximum of a twice differentiable function, the gradient vanishes there and the Hessian is negative semidefinite.

\emph{Part \rev{(a)}.}
Taylor's theorem at $\xi^{\ast}$ gives
\[
G(\xi^{\ast})-G(\xi)=\tfrac12(\xi-\xi^{\ast})^\top H(\xi-\xi^{\ast})+\rho(\xi),\qquad|\rho(\xi)|\le\epsilon\big(\lVert\xi-\xi^{\ast}\rVert\big)\,\lVert\xi-\xi^{\ast}\rVert^2,
\]
for a function $\epsilon$ with $\epsilon(t)\to0$ as $t\to0$.
By Assumption~\ref{ass:est}, $\sqrt N(\hat\xi_N-\xi^{\ast})$ converges in distribution to $Z$, so $\lVert\hat\xi_N-\xi^{\ast}\rVert\to0$ in probability and $N\rho(\hat\xi_N)\to0$ in probability.
The continuous mapping theorem gives
\[
\tfrac N2(\hat\xi_N-\xi^{\ast})^\top H(\hat\xi_N-\xi^{\ast})\to\tfrac12Z^\top HZ\quad\text{in distribution},
\]
and Slutsky's lemma adds the remainder.
The mean of the limit is
\[
\tfrac12\mathbb{E}[Z^\top HZ]=\tfrac12\operatorname{tr}\big(H\mathcal{I}(\xi^{\ast})^{-1}\big).
\]

\emph{Part \rev{(b)}.}
Let $\epsilon_f=G(\xi^{\ast})-\mathrm{Acc}_d(f)>0$.
The event $G(\hat\xi_N)\le\mathrm{Acc}_d(f)$ is the event
\[
N\big(G(\xi^{\ast})-G(\hat\xi_N)\big)\ge N\epsilon_f.
\]
The left side converges in distribution by part \rev{(a)}, so it is bounded in probability, whereas $N\epsilon_f\to\infty$.
The probability of the event therefore tends to zero.

\paragraph{Proof of Corollary~\ref{cor:nc}.}
For a parameter $\xi'$ that generates the ratings, let $G_{\xi'}(\xi)$ denote the accuracy of $m_\xi$ under the distribution $p_{\xi'}$.
The smoothness argument above applies jointly in $(\xi',\xi)$, so the Hessian
\[
H(\xi')=-\nabla^2_\xi G_{\xi'}(\xi)\big|_{\xi=\xi'}
\]
and the Fisher information $\mathcal{I}(\xi')$ are continuous in $\xi'$.
The information is nonsingular at $\xi^{\ast}$, so its determinant is positive on a neighborhood of $\xi^{\ast}$, and its inverse is continuous there.
Hence
\[
\kappa(\xi')=\tfrac12\operatorname{tr}\big(H(\xi')\mathcal{I}(\xi')^{-1}\big)
\]
is continuous at $\xi^{\ast}$.
The gap
\[
\Delta^{\mathrm{mean}}_d(\xi')=G_{\xi'}(\xi')-\mathrm{Acc}_{d,\xi'}(\text{mean})
\]
is a continuous function of the finite vectors $(p_{\xi'}(y))_y$ and $(m_{\xi'}(y))_y$, so it is continuous as well.
Assumption~\ref{ass:est} implies $\hat\xi_N\to\xi^{\ast}$ in probability, and the continuous mapping theorem gives the convergence of $\kappa(\hat\xi_N)$ and $\Delta^{\mathrm{mean}}_d(\hat\xi_N)$.
Since $\Delta^{\mathrm{mean}}_d(\xi^{\ast})>0$, the ratio $\kappa(\hat\xi_N)/\Delta^{\mathrm{mean}}_d(\hat\xi_N)$ converges in probability to $N_c$.

\subsection{The one-factor model}

\label{app:onefactor}
\rev{The gap $\Delta^{\mathrm{mean}}_d$ of \S\ref{sec:cost} has no closed form under the graded-response model, so we study it in the one-factor model of classical test theory \citep{lordnovick1968}.
In this model, the standardized rating of judge $i$ on item $j$ is $z_{ij}=r_i\theta_j+e_{ij}$, where $\theta_j$ has unit variance and $e_{ij}$ is independent normal noise with variance $1-r_i^2$.
The reliability $r_i\in(0,1)$ is thus the correlation of the ratings of judge $i$ with the latent quality, and it increases with the discrimination of the matching normal-ogive graded-response model \citep{takane1987}.
The posterior mean weights judge $i$ in proportion to $r_i/(1-r_i^2)$, whereas the mean gives every judge the same weight.
Let $\NSR(f)=(1-\mathrm{Acc}_d(f))/\mathrm{Acc}_d(f)$ be the noise-to-signal ratio of an aggregator.
Proposition~\ref{prop:gap} shows that the gap of the mean grows with the square of the reliability variation:}
\[
\frac{\NSR(\mathrm{mean})}{\NSR(m^{\ast})}=1+\Big(\frac{1+\bar r^2}{1-\bar r^2}\Big)^2\delta_r^2+O(\delta_r^3).
\]
\rev{For example, five judges with reliabilities $0.4,0.5,\ldots,0.8$ have $\delta_r=0.236$; the ceiling is $0.793$, and the mean reaches $0.744$.
In this example, the winning probability of \rev{Theorem~\ref{thm:region}} is $0.5$ at $23$ items and $0.9$ at $53$ items.}

A linear aggregator chooses a weight $w_i$ for each judge and returns $f_w(Y_j)=\sum_i w_iz_{ij}$.
The mean of the standardized ratings is the linear aggregator with $w_i=1/J$ for every judge.
Write $\psi_i^2=1-r_i^2$ for the noise variance of judge $i$.
Linearity gives $\operatorname{Cov}(f_w,\theta)=\sum_iw_ir_i$, and independence of the noise gives
\[
\operatorname{Var}f_w=\Big(\sum_iw_ir_i\Big)^2+\sum_iw_i^2\psi_i^2.
\]
Since $\operatorname{Var}\theta=1$, the accuracy is $\mathrm{Acc}(f_w)=1/(1+\NSR(f_w))$ with
\begin{equation}
\NSR(f_w)=\frac{\sum_iw_i^2\psi_i^2}{\big(\sum_iw_ir_i\big)^2},
\label{eq:corrnsr}
\end{equation}
which agrees with the definition in \S\ref{sec:synthetic}.
Let $\varphi_i=r_i^2/\psi_i^2$ denote the signal-to-noise ratio of judge $i$, and let $w_i^{\ast}=r_i/\psi_i^2$.
The latent quality and the ratings are jointly normal, so the posterior mean is linear in the ratings and proportional to $f_{w^{\ast}}$.
By Proposition~\ref{prop:ceiling}, whose proof applies to continuous ratings, $f_{w^{\ast}}$ attains the ceiling.
For numbers $x_i$, write $\CV_\varphi(x)$ for their $\varphi$-weighted coefficient of variation, the weighted standard deviation divided by the weighted mean.

\paragraph{\rev{Formal statements.}}
\rev{The analysis of \S\ref{sec:cost} in the one-factor model uses the following results.}

\begin{proposition}[Gap of the mean]
\label{prop:gap}
\rev{In the one-factor model, the following hold.}
\begin{enumerate}[label=(\alph*),leftmargin=*,itemsep=0pt,topsep=2pt]
  \item \rev{Every weighted sum $f_w=\sum_iw_iz_{ij}$ with $\sum_iw_ir_i\neq0$ satisfies
  \[
  \frac{\NSR(f_w)}{\NSR(m^{\ast})}=1+\CV_\varphi^2(w/w^{\ast}),
  \]
  where $w/w^{\ast}$ has entries $w_i/w_i^{\ast}$.}
  \item Let the mean average the standardized ratings.
  As $\delta_r\to0$,
  \[
  \frac{\NSR(\mathrm{mean})}{\NSR(m^{\ast})}=1+\Big(\frac{1+\bar r^2}{1-\bar r^2}\Big)^2\delta_r^2+O(\delta_r^3).
  \]
\end{enumerate}
\end{proposition}

\begin{lemma}[\rev{Local limits}]
\label{lem:region}
\rev{In the setting of Theorem~\ref{thm:region}, write the reliability of judge $i$ as $r_i=\bar r\,(1+u_i/\sqrt N)$ for fixed $u_i$ with $\sum_iu_i=0$ and $\frac1J\sum_iu_i^2=h^2$, so that $\delta_r=h/\sqrt N$.
As $N\to\infty$, the following hold.}
\begin{enumerate}[label=(\alph*),leftmargin=*,itemsep=0pt,topsep=2pt]
  \item \rev{$N$ times the loss $G(\xi^{\ast})-G(\hat\xi_N)$ of \MODEL{} converges in distribution to $\kappa_0/(J-1)$ times a chi-square variable with $J-1$ degrees of freedom.}
  \item \rev{$N\Delta^{\mathrm{mean}}_d\to\tau h^2$.}
\end{enumerate}
\rev{Here $\tau>0$ sets how fast the gap of the mean grows with the variation, and $\kappa_0>0$ is the estimation cost at equal reliabilities; both depend only on $J$ and $\bar r$ (Eq.~\eqref{eq:consts}).}
\end{lemma}
\rev{On $4{,}000$ simulated data sets of the five-judge example of App.~\ref{app:onefactor}, \MODEL{} is more accurate than the mean in $47\%$ of them at $N=25$ and in $96\%$ at $N=100$, against $55\%$ and $99\%$ from the limit of \rev{Theorem~\ref{thm:region}}.}

\begin{lemma}[Loss of a weighting]
\label{lem:loss}
In the one-factor model, $\NSR(f_{w^{\ast}})=1/\sum_i\varphi_i$.
For every weight vector $w$ with $\sum_iw_ir_i\neq0$, the ratio $q_i=w_i/w_i^{\ast}$ satisfies
\[
\frac{\NSR(f_w)}{\NSR(f_{w^{\ast}})}=1+\CV_\varphi^2(q).
\]
\end{lemma}

\begin{proof}
Write $w_i=q_iw_i^{\ast}=q_ir_i/\psi_i^2$.
Then $w_i^2\psi_i^2=q_i^2\varphi_i$ and $w_ir_i=q_i\varphi_i$, so
\[
\NSR(f_w)=\frac{\sum_i\varphi_iq_i^2}{\big(\sum_i\varphi_iq_i\big)^2}.
\]
With $q_i=1$ for every judge this gives $\NSR(f_{w^{\ast}})=1/\sum_i\varphi_i$.
The ratio of the two is
\[
\frac{\NSR(f_w)}{\NSR(f_{w^{\ast}})}=\frac{\mathbb{E}_\varphi q^2}{(\mathbb{E}_\varphi q)^2}=1+\CV_\varphi^2(q),
\]
where $\mathbb{E}_\varphi$ is the $\varphi$-weighted mean, and the last equality is the definition of the weighted coefficient of variation.
\end{proof}

A weighting loses nothing exactly when it is proportional to $w^{\ast}$.
The identity has the same form as the design effect of unequal weights in survey sampling \citep{kish1965}, the factor by which unequal weights inflate the variance of a weighted estimate.
In the five-judge example, the mean has $\NSR=0.344$ and the optimal weights have $\NSR=0.261$, a ratio of $1.317$\rev{, against $1.251$ from Proposition~\ref{prop:gap}(b)}.

\paragraph{Proof of Proposition~\ref{prop:gap}.}
\rev{Part (a) is Lemma~\ref{lem:loss}, because $m^{\ast}$ is a positive multiple of $f_{w^{\ast}}$ and has the same noise-to-signal ratio.
For part (b), the} ratio $\NSR(f_w)/\NSR(f_{w^{\ast}})$ does not change when $w$ is multiplied by a constant, so we may take $w_i=1$.
For the mean, $q_i=1/w_i^{\ast}=g(r_i)$ with $g(r)=(1-r^2)/r$.
The elasticity of $g$ is
\[
\frac{d\log g}{d\log r}=-\frac{1+r^2}{1-r^2}.
\]
Write $r_i=\bar r(1+u_i)$ with $\sum_iu_i=0$ and $\frac1J\sum_iu_i^2=\delta_r^2$.
A first-order expansion gives
\[
\log q_i=\log g(\bar r)-\frac{1+\bar r^2}{1-\bar r^2}\,u_i+O(u_i^2).
\]
For small spread, the squared coefficient of variation of $q$ equals the variance of $\log q$ up to third-order terms, and the $\varphi$-weighted variance of $u$ equals $\delta_r^2(1+O(\delta_r))$.
Lemma~\ref{lem:loss} then gives the result.

\paragraph{\rev{Proof of Lemma~\ref{lem:region} and Theorem~\ref{thm:region}.}}
\rev{The constants of \rev{Theorem~\ref{thm:region}} are}
\begin{equation}
A=\frac{J\bar r^2}{1+(J-1)\bar r^2},\qquad
\tau=A(1-A)\Big(\frac{1+\bar r^2}{1-\bar r^2}\Big)^2,\qquad
\kappa_0=\frac{(J-1)(1-\bar r^4)}{(J-2)\,\bar r^2\big(1+(J-1)\bar r^2\big)},
\label{eq:consts}
\end{equation}
\rev{where $A$ is the ceiling when every reliability equals $\bar r$.}

\rev{\emph{Theorem~\ref{thm:cost} in the one-factor model.}
The parameter is $\xi=(\lambda,\psi^2)\in\mathbb{R}^{2J}$, the loadings and noise variances of the factor model $z_j\sim\mathcal{N}\big(0,\lambda\lambda^\top+\operatorname{diag}(\psi^2)\big)$, with true value $\lambda_i=r_i$ and $\psi_i^2=1-r_i^2$.
With $J\ge3$ nonzero loadings, the model is identified up to the sign of $\lambda$, its Fisher information is nonsingular, and the maximum-likelihood estimate is asymptotically normal \citep{anderson1956}.
Assumption~\ref{ass:est} therefore holds with the sign fixed by $\lambda_1>0$.
The posterior mean under $\xi$ is a positive multiple of $f_{w(\xi)}$ with $w_i(\xi)=\lambda_i/\psi_i^2$, so $G(\xi)=\mathrm{Acc}(f_{w(\xi)})$ is a rational function of $\xi$ and is smooth near the true value.
The proof of Theorem~\ref{thm:cost} uses the finite rating set only to show that $G$ is twice continuously differentiable, so \rev{Lemma~\ref{lem:gmax} and part (a) of the theorem} hold.
The constant $\kappa$ is a smooth function of $r$ near $\bar r\mathbf{1}$, so $\kappa=\kappa_0+O(\delta_r)$, where $\kappa_0$ is its value at $r_i=\bar r$.}

\rev{\emph{Local parameters.}
Let $\xi_N$ be the true parameter at the reliabilities of \rev{the lemma}, and let $\xi_0$ be its value at $r=\bar r\mathbf{1}$, so $\sqrt N(\xi_N-\xi_0)$ converges.
The factor model is a smooth normal family with nonsingular information at $\xi_0$, so it is locally asymptotically normal, and the maximum-likelihood estimate is asymptotically linear in the score at $\xi_0$.
Le Cam's third lemma then gives $\sqrt N(\hat\xi_N-\xi_N)\to\mathcal{N}\big(0,\mathcal{I}(\xi_0)^{-1}\big)$ in distribution under $\xi_N$ \citep[Ch.~6--8]{vandervaart1998}.
Let $G_N(\xi)$ be the accuracy of $f_{w(\xi)}$ under $\xi_N$, and let $H_0=-\nabla^2G(\xi_0)$.
The function $G_N$ is maximized at $\xi_N$ with zero gradient, and its Hessian at $\xi_N$ converges to $-H_0$.
A second-order expansion therefore gives $N\big(G_N(\xi_N)-G_N(\hat\xi_N)\big)\to\frac12Z^\top H_0Z$ in distribution, with $Z\sim\mathcal{N}\big(0,\mathcal{I}(\xi_0)^{-1}\big)$.}

\rev{\emph{Gap.}
Since $\mathrm{Acc}=1/(1+\NSR)$,
\[
\Delta^{\mathrm{mean}}_d=\frac{\NSR(\mathrm{mean})-\NSR(m^{\ast})}{\big(1+\NSR(m^{\ast})\big)\big(1+\NSR(\mathrm{mean})\big)}.
\]
Proposition~\ref{prop:gap} gives $\NSR(\mathrm{mean})-\NSR(m^{\ast})=a^2\,\NSR(m^{\ast})\,\delta_r^2+O(\delta_r^3)$ with $a=(1+\bar r^2)/(1-\bar r^2)$.
By Lemma~\ref{lem:loss}, $\NSR(m^{\ast})=1/\sum_i\varphi_i=n_0+O(\delta_r)$ with $n_0=(1-\bar r^2)/(J\bar r^2)$.
Hence $\Delta^{\mathrm{mean}}_d=a^2n_0\delta_r^2/(1+n_0)^2+O(\delta_r^3)=\tau\delta_r^2+O(\delta_r^3)$, because $n_0/(1+n_0)^2=A(1-A)$ with $A=1/(1+n_0)$.}
\rev{With $\delta_r=h/\sqrt N$, this gives $N\Delta^{\mathrm{mean}}_d=\tau h^2+O(N^{-1/2})$, which is part (b).}

\rev{\emph{Estimation cost at equal reliabilities.}
Let $r_i=\bar r$ for every judge, write $\psi^2=1-\bar r^2$, and let $\Sigma=rr^\top+\psi^2I$ be the covariance of the standardized ratings.
For a weight vector $w$, $\mathrm{Acc}(f_w)=(w^\top r)^2/(w^\top\Sigma w)$, so
\[
A-\mathrm{Acc}(f_w)=\frac{w^\top(A\Sigma-rr^\top)\,w}{w^\top\Sigma w}.
\]
Direct computation gives $A\Sigma-rr^\top=A\psi^2P$ with $P=I-\mathbf{1}\mathbf{1}^\top/J$.
Since $Pw^{\ast}=0$, the numerator equals $A\psi^2\lVert P(w-w^{\ast})\rVert^2$.
Permutations of the judges leave the model unchanged, so the Fisher information does not mix perturbations along $\mathbf{1}$ with perturbations along a unit vector $e\perp\mathbf{1}$, nor two such vectors that are orthogonal.
For the perturbation $\xi=\xi^{\ast}+(s\,e,\,t\,e)$, the formula $\mathcal{I}_{ab}=\frac12\operatorname{tr}(\Sigma^{-1}\partial_a\Sigma\,\Sigma^{-1}\partial_b\Sigma)$ gives the information
\[
\mathcal{I}_e=\frac{1}{\psi^2(\psi^2+J\bar r^2)}\begin{pmatrix}J\bar r^2&\bar r\\[2pt]\bar r&\dfrac{\psi^2+(J-2)\bar r^2}{2\psi^2}\end{pmatrix},
\]
whose determinant is positive exactly when $J\ge3$.
Along this perturbation, $P(w-w^{\ast})=\zeta^\top(s,t)^\top e$ to first order, with $\zeta=(1/\psi^2,\,-\bar r/\psi^4)^\top$.
Summing over the $J-1$ orthogonal directions gives
\[
\kappa_0=(J-1)\,\frac{A\psi^2}{w^{\ast\top}\Sigma w^{\ast}}\,\zeta^\top\mathcal{I}_e^{-1}\zeta.
\]
With $w^{\ast}=(\bar r/\psi^2)\mathbf{1}$, the prefactor equals $\psi^6/\big(1+(J-1)\bar r^2\big)^2$.
Inverting the $2\times2$ matrix gives
\[
\zeta^\top\mathcal{I}_e^{-1}\zeta=\frac{(1+\bar r^2)\big(1+(J-1)\bar r^2\big)}{(J-2)\,\bar r^2\psi^4},
\]
and the product is the stated $\kappa_0$.}
\rev{The same block structure gives the law of $\frac12Z^\top H_0Z$.
Fix orthonormal vectors $e_1,\ldots,e_{J-1}\perp\mathbf{1}$, and let $Z_k$ be the pair of components of $Z$ along $e_k$ in the loadings and in the noise variances.
The pairs $Z_k$ are independent with covariance $\mathcal{I}_e^{-1}$, and the components of $Z$ along $\mathbf{1}$ leave $P(w-w^{\ast})$ unchanged to first order.
Hence $\frac12Z^\top H_0Z=\frac{A\psi^2}{w^{\ast\top}\Sigma w^{\ast}}\sum_{k=1}^{J-1}(\zeta^\top Z_k)^2$, a sum of $J-1$ independent terms with the same scale.
Each $\zeta^\top Z_k$ is normal with variance $\zeta^\top\mathcal{I}_e^{-1}\zeta$, so the sum is $\kappa_0/(J-1)$ times a chi-square variable with $J-1$ degrees of freedom, which is part (a).}

\rev{\emph{Winning probability.}
\MODEL{} is more accurate than the mean exactly when $N\big(G_N(\xi_N)-G_N(\hat\xi_N)\big)<N\Delta^{\mathrm{mean}}_d$.
The limit law of part (a) is continuous, so parts (a) and (b) and Slutsky's lemma give \rev{Theorem~\ref{thm:region}}.}

\rev{\emph{Fixed variation.}
For fixed reliabilities, dividing the two expansions gives $N_c=\kappa/\Delta^{\mathrm{mean}}_d=\big(\kappa_0+O(\delta_r)\big)/\big(\tau\delta_r^2+O(\delta_r^3)\big)=\kappa_0\big(1+O(\delta_r)\big)/(\tau\delta_r^2)$.
Together with the expansion of the gap above, this proves Proposition~\ref{prop:nc}.}

\begin{lemma}[Label-free \rev{variation}]
\label{lem:delta}
In the one-factor model with population correlations, let $\bar\rho_i$ be the mean correlation of judge $i$ with the other judges, and let $\delta$ be the coefficient of variation of $\bar\rho_1,\ldots,\bar\rho_J$. As $\delta_r\to0$,
\[
\delta=\frac{J-2}{J-1}\,\delta_r+O(\delta_r^2).
\]
\end{lemma}

\begin{proof}
The population correlation of judges $i$ and $k$ is $r_ir_k$.
The mean correlation of judge $i$ with the others is therefore
\[
\bar\rho_i=\frac{r_i(J\bar r-r_i)}{J-1}=\bar r^2\Big(1+\frac{J-2}{J-1}\,u_i+O(u_i^2)\Big),
\]
where the second form uses $r_i=\bar r(1+u_i)$.
The coefficient of variation of $\bar\rho_i$ is therefore $\frac{J-2}{J-1}\delta_r+O(\delta_r^2)$.
\end{proof}

A judge's mean agreement with the others is proportional to its own reliability, up to a correction that vanishes as $J$ grows.
In the five-judge example, the factor is $3/4$, and the agreement \rev{variation} is $0.180$ against the predicted $0.177$.

\subsection{Dawid--Skene}
\label{app:ds}

\paragraph{\rev{Formal statements.}}
Dawid--Skene treats the quality of an item as a class $c\in\{0,\ldots,K-1\}$, one class per rating category, with probabilities $\pi_c$.
Given class $c$, judge $i$ gives rating $k$ with probability $C_i(c,k)$, so each judge has a confusion \rev{matrix} with $K(K-1)$ free probabilities.
Dawid--Skene fits $\eta=(\pi,C)$ by maximum likelihood and scores item $j$ by its expected class $r_{\hat\eta}(Y_j)=\mathbb{E}_{\hat\eta}[c\mid Y_j]$.

Under Assumption~\ref{ass:klevel}, the model of Dawid--Skene is correct, because each judge has a valid confusion \rev{matrix}, and Theorem~\ref{thm:ds} compares the two methods.

\paragraph{\revred{Continuous quality.}}

Under Assumption~\ref{ass:model}, the latent quality is continuous, so the class model of Dawid--Skene is misspecified.
Its estimate then converges to the pseudo-true value $\eta^{\dagger}$, the parameter of the class model closest in Kullback--Leibler divergence to the true distribution of the ratings \citep{white1982}.
\rev{Let $\hat\eta_N$ be the maximum-likelihood estimate of $\eta$ from $N$ items. We assume the standard behavior of a misspecified maximum-likelihood estimate:}
\begin{assumption}[\rev{Regular misspecified estimation}]
\label{ass:dsest}
\rev{The expected log-likelihood of the class model has a unique interior maximizer $\eta^{\dagger}$ with a nonsingular Hessian, $\sqrt N(\hat\eta_N-\eta^{\dagger})$ converges in distribution to a normal law, and $\mathrm{Acc}_d(r_\eta)$ is differentiable in $\eta$ at $\eta^{\dagger}$.}
\end{assumption}
The loss of the limiting score is the floor
\[
\Delta^{\mathrm{DS}}_d=\mathrm{Acc}_d(m^{\ast})-\mathrm{Acc}_d(r_{\eta^{\dagger}}).
\]

\begin{proposition}[Floor of Dawid--Skene]
\label{prop:dsfloor}
Let Assumption~\ref{ass:model} hold with one dimension.
Let $\Omega$ be a connected open set of judge parameters on which $\eta^{\dagger}$ is the unique interior maximizer of the expected log-likelihood, its Hessian is nonsingular, and $\operatorname{Var}r_{\eta^{\dagger}}(Y_j)>0$.
\begin{enumerate}[label=(\alph*),leftmargin=*,itemsep=0pt,topsep=2pt]
  \item On $\Omega$, the loss $\mathrm{Acc}_d(m^{\ast})-\mathrm{Acc}_d(r_{\hat\eta_N})$ converges in probability to $\Delta^{\mathrm{DS}}_d$.
  \item If $\Delta^{\mathrm{DS}}_d>0$ at one point of $\Omega$, then $\Delta^{\mathrm{DS}}_d>0$ at almost every point of $\Omega$.
\end{enumerate}
\end{proposition}

\begin{theorem}[\rev{\MODEL{} versus Dawid--Skene}]
\label{thm:dsgap}
\rev{Let Assumptions~\ref{ass:model}, \ref{ass:est}, and~\ref{ass:dsest} hold with one dimension.}
\rev{If $\Delta^{\mathrm{DS}}_d>0$, then $P\big(G(\hat\xi_N)>\mathrm{Acc}_d(r_{\hat\eta_N})\big)\to1$.}
\end{theorem}
Proposition~\ref{prop:dsmain} combines Proposition~\ref{prop:dsfloor}(a) and Theorem~\ref{thm:dsgap}, with $\hat f^{\mathrm{DS}}_N=r_{\hat\eta_N}$.

{The floor is the limit of the loss. For finite $N$, the loss of Dawid--Skene also contains the cost of estimating $\eta$.
Let $\Gamma(\eta)=\mathrm{Acc}_d(r_\eta)$, so that $\Delta^{\mathrm{DS}}_d=G^{\ast}-\Gamma(\eta^{\dagger})$ with $G^{\ast}=G(\xi^{\ast})$.}
\rev{Let $q_\eta$ be the distribution of the rating patterns under the class model, $\ell(\eta)$ the expected log-likelihood, $A=-\nabla^2\ell(\eta^{\dagger})$, and $\mathbf{B}$ the covariance of $\nabla_\eta\log q_{\eta^{\dagger}}(Y_j)$.
Under Assumption~\ref{ass:dsest}, the normal limit has covariance $V=A^{-1}\mathbf{B}A^{-1}$ \citep{white1982}, and we write $g=\nabla\Gamma(\eta^{\dagger})$.}
{
\begin{lemma}[\rev{Fluctuation of the difference}]
\label{lem:dsclt}
\rev{Under the assumptions of Theorem~\ref{thm:dsgap},}
\[
\sqrt N\big(G(\hat\xi_N)-\Gamma(\hat\eta_N)-\Delta^{\mathrm{DS}}_d\big)\to\mathcal{N}(0,g^\top Vg)\quad\text{in distribution.}
\]
\end{lemma}
\begin{proposition}[\rev{Expected difference}]
\label{prop:dsexp}
\rev{Under the assumptions of Theorem~\ref{thm:dsgap}, suppose $\mathbb{E}[G^{\ast}-G(\hat\xi_N)]=\kappa/N+o(1/N)$ and $\mathbb{E}[G^{\ast}-\Gamma(\hat\eta_N)]=\Delta^{\mathrm{DS}}_d+\kappa_{\mathrm{DS}}/N+o(1/N)$. Then}
\[
\mathbb{E}\big[G(\hat\xi_N)-\Gamma(\hat\eta_N)\big]=\Delta^{\mathrm{DS}}_d+\frac{\kappa_{\mathrm{DS}}-\kappa}{N}+o\Big(\frac1N\Big).
\]
\end{proposition}
\revred{The first term does not shrink with $N$, so on large sets \MODEL{} stays ahead of Dawid--Skene by the floor; the second term is the difference in the cost of estimating the parameters, which matters most on small sets.}
The fluctuations of the two methods have different orders.
The true parameters maximize $G$, so the fluctuation of \MODEL{} is of order $1/N$ (Lemma~\ref{lem:gmax}).
The pseudo-true value $\eta^{\dagger}$ maximizes the likelihood, not $\Gamma$, so $g\neq0$ in general and the fluctuation of Dawid--Skene is of order $N^{-1/2}$.
Under Assumption~\ref{ass:klevel}, $\Delta^{\mathrm{DS}}_d=0$ and Theorem~\ref{thm:ds} gives the constants $\kappa_K$ and $\kappa_{\mathrm{DS}}$.}

\rev{\emph{Comparison with the mean.}
Dawid--Skene becomes more accurate than the mean with probability tending to one only if $\Delta^{\mathrm{mean}}_d\ge \Delta^{\mathrm{DS}}_d$, whereas \MODEL{} needs only $\Delta^{\mathrm{mean}}_d>0$ (Theorem~\ref{thm:cost}(b)).
In both settings, \MODEL{} (\MODEL{}-$K$ with $K$ levels) becomes more accurate than the mean wherever Dawid--Skene does, and with $K$ levels it needs no more items to do so.}

\paragraph{Notation.}
Fix one dimension and drop the index $d$.
The rating vector $Y_j$ takes values in a finite set $\mathcal{Y}$, and every pattern has positive probability.
A score is a vector $t\in\mathbb{R}^{\mathcal{Y}}$, and all moments are under the true distribution.
Accuracy is invariant to location and positive scale of the latent quality, so we take $\operatorname{Var}\theta_j=1$ under both Assumption~\ref{ass:model} and Assumption~\ref{ass:klevel}.
Write $G^{\ast}=\mathrm{Acc}(m^{\ast})=\operatorname{Var}m^{\ast}(Y_j)$, which holds under both assumptions by Proposition~\ref{prop:ceiling}.

\begin{lemma}[Local form of the loss]
\label{lem:local}
For every $u\in\mathbb{R}^{\mathcal{Y}}$ with $\operatorname{Var}(m^{\ast}+u)>0$,
\[
G^{\ast}-\mathrm{Acc}(m^{\ast}+u)=\frac{G^{\ast}\,u^\top Qu}{\operatorname{Var}\big(m^{\ast}(Y_j)+u(Y_j)\big)},\qquad u^\top Qu=\operatorname{Var}u-\frac{\operatorname{Cov}(u,m^{\ast})^2}{G^{\ast}}.
\]
The matrix $Q$ is positive semidefinite, and $Q\mathbf{1}=Qm^{\ast}=0$.
\end{lemma}

\begin{proof}
The tower property gives $\operatorname{Cov}(t,\theta_j)=\operatorname{Cov}(t,m^{\ast})$ for every score $t$, and $\operatorname{Var}\theta_j=1$, so $\mathrm{Acc}(t)=G^{\ast}\operatorname{corr}^2(t,m^{\ast})$.
For $t=m^{\ast}+u$, the residual of $t$ after linear regression on $m^{\ast}$ is the residual of $u$, so $1-\operatorname{corr}^2(t,m^{\ast})=u^\top Qu/\operatorname{Var}t$.
The form $u^\top Qu$ is a residual variance, so it is nonnegative, and it vanishes for constants and for multiples of $m^{\ast}$.
\end{proof}

Lemma~\ref{lem:local} gives $H=2\nabla m^\top Q\nabla m$ in Theorem~\ref{thm:cost}, where $\nabla m$ is the Jacobian of $\xi\mapsto m_\xi$ at $\xi^{\ast}$.
The next lemma shows that no regular estimator of the ceiling score has a smaller loss constant than $\kappa$.
An estimator is regular if its limit law does not change under small perturbations of the true parameter \citep[Ch.~8]{vandervaart1998}.

\begin{lemma}[Efficiency of $\kappa$]
\label{lem:eff}
Let the ratings follow Assumption~\ref{ass:model} or Assumption~\ref{ass:klevel}, let $\xi$ be the parameter of \MODEL{} or of \MODEL{}-$K$, and let Assumption~\ref{ass:est} hold.
Let $T_N\in\mathbb{R}^{\mathcal{Y}}$ be a regular estimator of the score $\mu(\xi)=a(\xi)\mathbf{1}+b(\xi)m_\xi$, where $a$ and $b$ are continuously differentiable and $b(\xi^{\ast})>0$.
Then $N\big(G^{\ast}-\mathrm{Acc}(T_N)\big)$ converges in distribution to a law with mean at least $\kappa$.
\end{lemma}

\begin{proof}
Write $T_N=a^{\ast}\mathbf{1}+b^{\ast}m^{\ast}+D_N$ with $a^{\ast}=a(\xi^{\ast})$ and $b^{\ast}=b(\xi^{\ast})$.
Accuracy is invariant to location and positive scale, so the loss equals $G^{\ast}-\mathrm{Acc}(m^{\ast}+D_N/b^{\ast})$.
Regularity gives $\sqrt ND_N\to L$ in distribution, and Lemma~\ref{lem:local} with the continuous mapping theorem gives $N$ times the loss $\to L^\top QL/b^{\ast2}$.
The convolution theorem \citep[Thm.~8.8]{vandervaart1998} gives $L=W+V$, where $W\sim\mathcal{N}(0,\nabla\mu\,\mathcal{I}^{-1}\nabla\mu^\top)$ is independent of $V$.
Independence and $\mathbb{E}W=0$ give
\[
\mathbb{E}[L^\top QL]=\mathbb{E}[W^\top QW]+\mathbb{E}[V^\top QV]\ge\operatorname{tr}\big(Q\nabla\mu\,\mathcal{I}^{-1}\nabla\mu^\top\big).
\]
Since $Q\mathbf{1}=Qm^{\ast}=0$, $Q\nabla\mu=b^{\ast}Q\nabla m$, so the bound equals $b^{\ast2}\operatorname{tr}(Q\nabla m\,\mathcal{I}^{-1}\nabla m^\top)=b^{\ast2}\kappa$.
Dividing by $b^{\ast2}$ gives a limit mean of at least $\kappa$.
\end{proof}

\paragraph{Proof of Theorem~\ref{thm:ds}.}
(a) The proof of Theorem~\ref{thm:cost} uses only the smoothness of the posterior mean in the parameters and Assumption~\ref{ass:est}.
Under Assumption~\ref{ass:klevel} the posterior mean of \MODEL{}-$K$ is a finite sum of smooth terms, so Theorem~\ref{thm:cost}\rev{(a)} holds with constant $\kappa_K$.

\revp{Next, label} the classes of Dawid--Skene in the order of the true levels, and let $v_0$ and $\ell$ be the lowest level and the spacing of the levels.
Under Assumption~\ref{ass:klevel}, the class score at the parameter $\eta$ that corresponds to $\xi$ is
\[
r_\eta(y)=\mathbb{E}_\eta[c\mid y]=\frac{m_\xi(y)-v_0}{\ell},
\]
which has the form $a+b\,m_\xi$ with $b=1/\ell>0$.
The model of \MODEL{}-$K$ is a submodel of Dawid--Skene, so every smooth path of \MODEL{}-$K$ is a smooth path of Dawid--Skene.
The maximum-likelihood estimate of Dawid--Skene is regular in its own correct model, so it is regular along these paths.
Lemma~\ref{lem:eff} applied to \MODEL{}-$K$ gives a limit mean of at least $\kappa_K$, and $\kappa_{\mathrm{DS}}-\kappa_K$ is the excess $\ell^2\,\mathbb{E}[V^\top QV]\ge0$.

\revp{(b)} For $K=2$, both models share the level probabilities, and each judge has two rating probabilities $p_c=P(Y_{ij}=1\mid\theta_j=v_c)$ at the two levels $v_0<v_1$.
The map $(\alpha_i,\beta_i)\mapsto(p_0,p_1)$ with $p_c=\sigma(\alpha_iv_c-\beta_i)$ is a smooth bijection onto the open set $\{p_1>p_0\}$, which contains the true value.
The two models therefore coincide near the true parameter, their estimates agree with probability tending to one, and their scores are affine in each other, so $\kappa_{\mathrm{DS}}=\kappa_K$.
For $3\le K\le10$, \revred{let $\Omega_K$ be the set of parameters that satisfy Assumption~\ref{ass:klevel}. It is a product of convex sets (level probabilities, discriminations, and ordered thresholds), so it is connected.
For each model, let $\mathcal{I}$ be the Fisher information of one item and $H$ the negative Hessian of the accuracy in its own parameters, so that its loss constant is $\tfrac12\operatorname{tr}(\mathcal{I}^{-1}H)$.
Every rating pattern has positive probability, so these matrices are real-analytic on $\Omega_K$, and so is
\[
F=\tfrac12\Big(\det\mathcal{I}_K\operatorname{tr}\big(\operatorname{adj}(\mathcal{I}_{\mathrm{DS}})H_{\mathrm{DS}}\big)-\det\mathcal{I}_{\mathrm{DS}}\operatorname{tr}\big(\operatorname{adj}(\mathcal{I}_K)H_K\big)\Big),
\]
which equals $\det\mathcal{I}_K\det\mathcal{I}_{\mathrm{DS}}\,(\kappa_{\mathrm{DS}}-\kappa_K)$ where both informations are nonsingular.
A real-analytic function on a connected open set that is not identically zero vanishes only on a set of Lebesgue measure zero.
At the parameters of Table~\ref{tab:dskw}, both informations are nonsingular and $\kappa_{\mathrm{DS}}>\kappa_K$ for every $K$ from $3$ to $10$, so $F$ is not identically zero.
Hence $F\neq0$ almost everywhere on $\Omega_K$, so $\kappa_{\mathrm{DS}}\neq\kappa_K$ almost everywhere, and (a) gives $\kappa_{\mathrm{DS}}>\kappa_K$.}

\paragraph{Proof of Proposition~\ref{prop:dsfloor}.}
(a) The maximum-likelihood estimate of a misspecified model converges in probability to the unique pseudo-true value $\eta^{\dagger}$ \citep{white1982}.
Every pattern has positive probability under $\eta^{\dagger}$, so $r_\eta$ is a continuous function of $\eta$ near $\eta^{\dagger}$.
Accuracy is a continuous function of the finite vector $r_\eta$ where its variance is positive, and $\operatorname{Var}r_{\eta^{\dagger}}>0$.
The continuous mapping theorem gives the convergence of the loss to $\Delta^{\mathrm{DS}}_d$.

(b) For each pattern $y$, $p_{\xi^{\ast}}(y)$ is a real-analytic function of $(\alpha^{\ast},\beta^{\ast})$, because it is a Gaussian integral of a product of logistic functions.
The pseudo-true value solves $\sum_yp_{\xi^{\ast}}(y)\nabla_\eta\log q_\eta(y)=0$, where $q_\eta$ is the pattern distribution of the class model.
This equation is analytic in $(\alpha^{\ast},\beta^{\ast},\eta)$ with a nonsingular derivative in $\eta$, so the analytic implicit function theorem makes $\eta^{\dagger}$ real-analytic on $\Omega$.
Hence $\Delta^{\mathrm{DS}}_d$ is real-analytic on $\Omega$, and $\Delta^{\mathrm{DS}}_d\ge0$ by Proposition~\ref{prop:ceiling}.
If $\Delta^{\mathrm{DS}}_d$ is positive at one point of the connected set $\Omega$, it is not identically zero, so it vanishes only on a set of Lebesgue measure zero.

{\paragraph{\rev{Proof of Lemma~\ref{lem:dsclt}, Theorem~\ref{thm:dsgap}, and Proposition~\ref{prop:dsexp}.}}
\rev{Lemma~\ref{lem:dsclt}.} By the definition of $\Delta^{\mathrm{DS}}_d$,
\[
G(\hat\xi_N)-\Gamma(\hat\eta_N)-\Delta^{\mathrm{DS}}_d=\big[G(\hat\xi_N)-G^{\ast}\big]-\big[\Gamma(\hat\eta_N)-\Gamma(\eta^{\dagger})\big].
\]
Theorem~\ref{thm:cost}(a) gives $N\big(G^{\ast}-G(\hat\xi_N)\big)=O_p(1)$, so $\sqrt N\big(G(\hat\xi_N)-G^{\ast}\big)\to0$ in probability.
Assumption~\ref{ass:dsest} and the delta method give $\sqrt N\big(\Gamma(\hat\eta_N)-\Gamma(\eta^{\dagger})\big)\to\mathcal{N}(0,g^\top Vg)$.
Slutsky's lemma combines the two terms.

\rev{Theorem~\ref{thm:dsgap}.} By Lemma~\ref{lem:dsclt}, $\sqrt N\big(G(\hat\xi_N)-\Gamma(\hat\eta_N)-\Delta^{\mathrm{DS}}_d\big)$ is tight, whereas $-\sqrt N \Delta^{\mathrm{DS}}_d\to-\infty$.
Hence $P\big(G(\hat\xi_N)-\Gamma(\hat\eta_N)>0\big)\to1$.

\rev{Proposition~\ref{prop:dsexp}.} Subtract the first expansion from the second.}

\paragraph{Numerical values.}
The synthetic parameters are those of \rev{App.~\ref{app:synth}}: $J=5$ judges, discriminations $1.5\exp(s\,u_i)$ with $u_i$ evenly spaced on $[-1,1]$, and the fixed leniencies of that study.
For Theorem~\ref{thm:ds}, the $K$ levels are evenly spaced on $[-2,2]$, and their probabilities are those of a standard normal quality rounded to the nearest level.
The constants are computed exactly over all rating patterns, and each simulated data set is fitted to convergence by both methods.
Table~\ref{tab:dsk} lists $N$ times the loss, in units of accuracy.
The Dawid--Skene losses lie below their constants at these sizes, because a confusion cell with small probability adds to the loss only after the data contain it.
\revred{Table~\ref{tab:dskw} lists the constants for $K$ from $3$ to $10$ at $s=0.7$, with thresholds evenly spaced on $[-1.5,1.5]$ plus the fixed leniencies.}
\rev{With spread $s=0.7$ and $K=3$, the constant is $\kappa_3=2.35$.
On $500$ simulated data sets for each $N$ from $2{,}000$ to $32{,}000$, $N$ times the loss is $2.32$--$2.36$ for \MODEL{}-$K$ and $6.5$--$8.1$ for Dawid--Skene, and the loss of Dawid--Skene is larger in more than $99\%$ of the data sets.}
For Proposition~\ref{prop:dsfloor}, the floor is computed exactly over all rating patterns with Dawid--Skene fitted to convergence.
\rev{The floor is $0.0023$ for $s=0$ and $0.0040$ for $s=0.7$, so Proposition~\ref{prop:dsfloor}(b) applies around these parameters.}
The floor grows with the number of judges (Table~\ref{tab:dsfloor}), where $J\ge10$ uses $10^5$ items for fitting and $2\cdot10^5$ items for evaluation.

\begin{table}[h]
\centering\small
\begin{tabular}{lccccc}
\toprule
& \multicolumn{2}{c}{constant} & \multicolumn{3}{c}{simulated, $N=2{,}000$ to $32{,}000$}\\
\cmidrule(lr){2-3}\cmidrule(lr){4-6}
Setting & \MODEL{}-$K$ & Dawid--Skene & \MODEL{}-$K$ & Dawid--Skene & Dawid--Skene larger\\
\midrule
$K=3$, $s=0.7$ & $2.35$ & $8.37$ & $2.32$--$2.36$ & $6.5$--$8.1$ & $99.4$--$99.8\%$\\
$K=5$, $s=0.7$ & $3.81$ & $45.7$ & $3.80$--$4.39$ & $34.6$--$45.5$ & $100\%$\\
$K=5$, $s=1.4$ & $3.04$ & $52.5$ & $3.03$--$3.60$ & $17.5$--$19.8$ & $99.6$--$100\%$\\
\bottomrule
\end{tabular}
\caption{Scaled loss $N\big(G^{\ast}-\mathrm{Acc}\big)$ under Assumption~\ref{ass:klevel}, $500$ data sets per $N$.}
\label{tab:dsk}
\end{table}

\begin{table}[h]
\centering\small
\revred{\begin{tabular}{lcccccccc}
\toprule
$K$ & $3$ & $4$ & $5$ & $6$ & $7$ & $8$ & $9$ & $10$\\
\midrule
$\kappa_K$ & $2.35$ & $3.03$ & $3.81$ & $4.74$ & $5.71$ & $6.82$ & $8.10$ & $9.52$\\
$\kappa_{\mathrm{DS}}$ & $8.37$ & $20.9$ & $45.7$ & $98.4$ & $314$ & $2.6\cdot10^3$ & $3.9\cdot10^4$ & $6.9\cdot10^5$\\
\bottomrule
\end{tabular}}
\caption{\revred{Loss constants of \MODEL{}-$K$ ($\kappa_K$) and Dawid--Skene ($\kappa_{\mathrm{DS}}$) at $s=0.7$, computed exactly over all rating patterns.}}
\label{tab:dskw}
\end{table}

\begin{table}[h]
\centering\small
\begin{tabular}{lcccc}
\toprule
Judges $J$ & $5$ & $10$ & $20$ & $40$\\
\midrule
$\Delta^{\mathrm{DS}}_d$, $s=0$ & $0.0023$ & $0.011$ & $0.030$ & $0.050$\\
$\Delta^{\mathrm{DS}}_d$, $s=0.7$ & $0.0040$ & $0.017$ & $0.037$ & $0.055$\\
\bottomrule
\end{tabular}
\caption{Floor of Dawid--Skene under Assumption~\ref{ass:model} with $K=5$ rating categories.}
\label{tab:dsfloor}
\end{table}

\section{Judge Pool and Prompts}
\label{app:prompts}
The candidate set has ten instruction-tuned models: three proprietary models (\texttt{gpt-5}, \texttt{gemini-2.5-pro}, \texttt{claude-sonnet-4-6}) and seven open-weight models (\texttt{DeepSeek-R1}, \texttt{Qwen3-Next-80B}, \texttt{Llama-4-Scout}, \texttt{gpt-oss-120b}, \texttt{gemma-4-31B}, \texttt{GLM-4.7-Flash}, \texttt{Nemotron-120B}).
For each benchmark and dimension, every judge scores one item at a time on the benchmark's integer scale and returns the score as JSON, decoded at temperature $0$.
No human labels are used to tune or calibrate the judges.
Table~\ref{tab:perjudge} reports each judge's accuracy against the \rev{human ratings} on the full benchmarks.
The prompts are shown below.

\begin{promptbox}[SummEval]
\footnotesize\ttfamily\setlength{\parindent}{0pt}
You are an expert evaluator assessing \{dimension\}.\\
Source Article: \{source\}\\
Summary: \{summary\}\\
Rate on 1-5 (5=Excellent ... 1=Very Poor).\\
Output ONLY valid JSON: \{"score": N\}
\end{promptbox}
\noindent\,\textrm{\small \{dimension\}: coherence, consistency, fluency,
relevance. The \texttt{Source Article} line is included only for consistency and
relevance.}

\begin{promptbox}[Topical-Chat]
\footnotesize\ttfamily\setlength{\parindent}{0pt}
You are an expert evaluator of open-domain dialog responses.\\
Assess \{dimension\}.\\
Dialog history: \{context\}\\
Background fact: \{fact\}\\
Candidate response: \{response\}\\
Rate on integer scale \{lo\}-\{hi\} (\{anchors\}).\\
Output ONLY valid JSON: \{"score": N\}
\end{promptbox}
\noindent\,\textrm{\small \{dimension\}: the six USR dimensions (Understandable
$0$--$1$; Natural, Maintains Context, Engaging each $1$--$3$; Uses Knowledge
$0$--$1$; Overall $1$--$5$); \{anchors\} gives that scale's endpoint labels.}

\begin{promptbox}[WMT22 en--de]
\footnotesize\ttfamily\setlength{\parindent}{0pt}
You are an expert evaluating machine translation quality.\\
Source (English): \{src\}\\
Translation (German): \{mt\}\\
Rate overall quality on 1-5 (5=Excellent ... 1=Very Poor).\\
Output ONLY valid JSON: \{"score": N\}
\end{promptbox}

\begin{table}[h]
\centering\small
\setlength{\tabcolsep}{4pt}
\begin{tabular}{lccc}
\toprule
Judge & SummEval & Topical-Chat & WMT22 \\
\midrule
gpt-5             & 0.530 & 0.629 & 0.427 \\
gemini-2.5-pro    & 0.550 & 0.655 & 0.421 \\
claude-sonnet-4-6 & 0.531 & 0.700 & 0.417 \\
\midrule
DeepSeek-R1       & 0.445 & 0.492 & 0.358 \\
Qwen3-Next-80B    & 0.495 & 0.567 & 0.358 \\
Llama-4-Scout     & 0.422 & 0.440 & 0.312 \\
gpt-oss-120b      & 0.480 & 0.584 & 0.368 \\
gemma-4-31B       & 0.566 & 0.704 & 0.432 \\
GLM-4.7-Flash     & 0.412 & 0.503 & 0.322 \\
Nemotron-120B     & 0.402 & 0.500 & 0.347 \\
\bottomrule
\end{tabular}
\caption{Spearman $\rho$ of each candidate judge against the \rev{human ratings} on the full benchmarks, averaged over dimensions.}
\label{tab:perjudge}
\end{table}

\section{Baseline Implementations}
\label{app:baselines}
All baselines see the same \rev{rating matrix} as \MODEL{} and use no human labels.
The \textbf{mean} and \textbf{median} average or take the median of the raw scores across judges.
The \textbf{trimmed mean} sorts each item's judge scores and discards $\lfloor 0.2J\rfloor$ scores from each end before averaging; at $J{=}5$ this removes the single highest- and lowest-scoring judge and averages the remaining three.
\textbf{Dawid--Skene} runs the EM algorithm of \citet{dawid1979} from majority-vote class probabilities until the log-likelihood gain per item falls below $10^{-9}$, so it returns the maximum-likelihood fit. \textbf{MACE} and \textbf{GLAD} are the \texttt{crowd-kit} library implementations: MACE runs $10$ iterations with one restart, and GLAD runs at most $10$ iterations under the library's default stopping rule, with five inner M-step iterations. Because these estimators are categorical, each integer rating level is treated as a separate class; we recover a scalar per item as the posterior expected rating $\sum_k k\,P(Y{=}k)$. This categorical treatment discards the ordinal structure that \MODEL{}'s graded-response link uses.
\textbf{CARE} \citep{zhao2026care} is run at its label-free library default (hyperparameter $\gamma{=}1$) with the SCS conic solver, taking the score-weighted quality vector as output.
Each aggregator returns one scalar per item, standardized before the Spearman comparison.

\section{\rev{Synthetic Data}}
\label{app:synth}
We draw ratings from a known graded-response model with $J=5$ judges and $K=5$ categories.
The discriminations are $\alpha_i=1.5\exp(s\,u_i)$ with $u_i$ evenly spaced on $[-1,1]$, so the spread parameter $s$ controls the reliability \rev{variation}.
\rev{The thresholds lie at $-1.5,-0.5,0.5,1.5$ on the latent scale, shifted for each judge by a fixed leniency drawn from $\mathcal{N}(0,0.3^2)$.}
Five values of $s$ give \rev{levels of reliability variation} $\delta_r$ from $0$ to $0.46$, and for each $N\in\{25,\ldots,1{,}600\}$ we draw $100$ data sets.
\rev{We compare \MODEL{} ($R{=}I$) and the mean with the posterior mean under the true parameters, which attains the ceiling.
The comparison uses the noise-to-signal ratio of \rev{\S\ref{sec:synthetic}}.}

\begin{figure}[h]
  \centering
  \includegraphics[width=0.6\columnwidth]{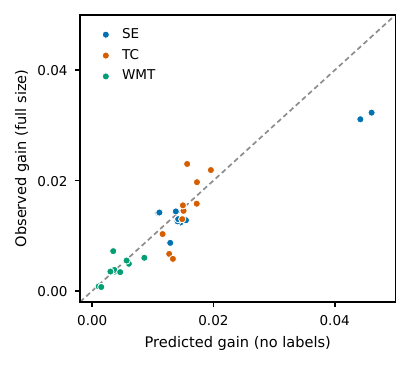}
  \caption{\rev{Ten real pools per benchmark: gain over the mean predicted without human labels, against the observed gain at full size. Dashed line: equality.}}
  \label{fig:synthreal}
\end{figure}

\paragraph{\rev{Label-free prediction on real pools.}}
\rev{Corollary~\ref{cor:nc} suggests that the gain over the mean and the break-even size can be predicted from the ratings alone.
To test this, we treat the \MODEL{} ($R{=}I$) fit of a real pool as the true model and simulate from it, without human labels.}
\rev{For each of the first $10$ pools of Table~\ref{tab:main}, we fit \MODEL{} ($R{=}I$) to all items of each dimension and treat the fit as the true model.
We draw rating sets of $N\in\{10,15,25,50,100\}$ items from this model and record the Spearman correlation of \MODEL{}, the mean, and the posterior mean under the fitted parameters with the simulated quality.
The predicted gain in \rev{Figure~\ref{fig:synthreal}} is the gain of the posterior mean over the mean at $N=100$, and the predicted break-even size is the set size at which the average simulated gain of \MODEL{} over the mean turns positive.
The observed gain is measured against the human ratings; human ratings that measure the latent quality with independent noise scale every correlation by the same factor (App.~\ref{app:proofs}), so the two gains are comparable.
At $25$ items, \MODEL{} exceeds the mean in $80\%$, $90\%$, and $40\%$ of these pools on SummEval, Topical-Chat, and WMT22, and in $90\%$ of the WMT22 pools at $50$ items.}
\rev{The predicted gain matches the observed gain on all three benchmarks and orders the pools, with Pearson correlations of $0.97$, $0.80$, and $0.71$ (Figure~\ref{fig:synthreal}).
The predicted break-even size lies between $10$ and $25$ items on all three benchmarks.}

\section{Uninformative Judges}
\label{app:noise}
We draw $10$ random pools of five judges and $10$ of seven judges from the candidate set, add $m$ judges whose ratings are drawn uniformly from the rating scale, and score $N=50$ items.
Table~\ref{tab:noise} reports the mean accuracy over the $20$ pools.
\rev{Without added judges ($m=0$), the mean agreement variation $\delta$ of these pools is $0.24$ on SummEval, $0.14$ on Topical-Chat, and $0.11$ on WMT22.
Over the $80$ combinations of pool and $m$ on each benchmark, the gain of \MODEL{} over the mean has a Spearman correlation with $\delta$ of $0.89$ on SummEval, $0.97$ on Topical-Chat, and $0.87$ on WMT22.}

\begin{table*}[t]
\centering\footnotesize
\setlength{\tabcolsep}{5pt}
\begin{tabular}{c ccc ccc ccc}
\toprule
& \multicolumn{3}{c}{\textbf{SummEval}} & \multicolumn{3}{c}{\textbf{Topical-Chat}} & \multicolumn{3}{c}{\textbf{WMT22}} \\
\cmidrule(lr){2-4}\cmidrule(lr){5-7}\cmidrule(lr){8-10}
$m$ & Mean & DS & \revp{\MODEL{}} & Mean & DS & \revp{\MODEL{}} & Mean & DS & \revp{\MODEL{}} \\
\midrule
  0 & 0.523 & 0.522 & \textbf{0.550} & 0.721 & 0.709 & \textbf{0.732} & 0.441 & 0.416 & \textbf{0.450}\\
  2 & 0.481 & 0.497 & \textbf{0.550} & 0.685 & 0.698 & \textbf{0.731} & 0.382 & 0.383 & \textbf{0.438}\\
  4 & 0.453 & 0.482 & \textbf{0.550} & 0.661 & 0.694 & \textbf{0.731} & 0.364 & 0.366 & \textbf{0.439}\\
  8 & 0.410 & 0.438 & \textbf{0.548} & 0.620 & 0.686 & \textbf{0.730} & 0.339 & 0.344 & \textbf{0.443}\\
\bottomrule
\end{tabular}
\caption{Spearman $\rho$ against the \rev{human ratings} at $N=50$ when $m$ uninformative judges join a pool of five or seven judges. DS: Dawid--Skene.}
\label{tab:noise}
\end{table*}

\end{document}